# A Bayesian Nonparametric Approach for Estimating Individualized Treatment-Response Curves

**Yanbo Xu** yxu68@jhu.edu
*Department of Computer Science, Johns Hopkins University*
*160 Malone Hall*
*3400 North Charles Street*
*Baltimore, MD 21218-2608*

**Yanxun Xu** yanxun.xu@jhu.edu
*Department of Applied Mathematics and Statistics, Johns Hopkins University*
*Whitehead Hall 100*
*3400 North Charles Street*
*Baltimore, MD 21218-2608*

**Suchi Saria** ssaria@cs.jhu.edu
*Department of Computer Science, Johns Hopkins University*
*160 Malone Hall*
*3400 North Charles Street*
*Baltimore, MD 21218-2608*

## Abstract

We study the problem of estimating the continuous response over time to interventions using *observational time series*—a retrospective dataset where the policy by which the data are generated is unknown to the learner. We are motivated by applications where response varies by individuals and therefore, estimating responses at the individual-level is valuable for personalizing decision-making. We refer to this as the problem of estimating individualized treatment response (ITR) curves. In statistics, G-computation formula (Robins, 1986) has been commonly used for estimating treatment responses from observational data containing sequential treatment assignments. However, past studies have focused predominantly on obtaining point-in-time estimates at the population level. We leverage the G-computation formula and develop a novel Bayesian nonparametric (BNP) method that can flexibly model functional data and provide posterior inference over the treatment response curves at both the individual and population level. On a challenging dataset containing time series from patients admitted to a hospital, we estimate responses to treatments used in managing kidney function and show that the resulting fits are more accurate than alternative approaches. Accurate methods for obtaining ITRs from observational data can dramatically accelerate the pace at which personalized treatment plans become possible.

## 1. Introduction

Accurate models of actions and their effects on the state of the agent are critical for decision-making. Learning of action-effect models is most straightforward from data where the learner can control the choice of actions and observe their responses. However, such data are not always possible to acquire. Alternatively, retrospective data may be available containing time series generated from observing the actions from other agents. Estimating action-effect models from *observational data*, where the learner cannot control the actions that are prescribed, or the actions may be prescribed by a mechanism that is not known to the learner, is more challenging. We study an instance of this problem: specifically, we consider the problem of estimating the continuous response over time to an action. We are particularly motivated by applications in medicine where accurate action-effect models for estimating treatment effects can be used for personalizing therapy.

In statistics, the problem of estimating treatment effects from observational data containing sequential treatment assignments has been studied extensively using approaches such as the G-computation formula (Robins, 1986), G-estimation of structural nested models (Robins, 2004), inverse probability of treatment weighted (IPTW) estimation of marginal structure models (van der Laan and Petersen, 2007), doubly robust learning (Tsiatis, 2007; Zhao et al., 2015) with applications to longitudinal data analysis (Hernán et al., 2000), survival analysis (Lunceford et al., 2002), and adaptive treatment selections in clinical trials (Murphy et al., 2007a,b). A related problem in reinforcement learning is off-policy evaluation where the goal is to estimate the value of a policy (sequence of actions) from data collected by another policy (Sutton et al., 1998). For example, doubly-robust estimators for policy evaluation have been developed for contextual bandits (Dudik et al., 2011) and for sequential decision-making problems (Jiang and Li, 2015). See survey of example techniques in Paduraru et al. (2012). In this paper, we use the G-computation formula to adjust for time-varying confounding. We depart from the existing literature by using a novel Bayesian nonparametric method to (1) flexibly model the longitudinal outcome over time, and (2) characterize heterogeneity in treatment effects across individuals. [1]

Bayesian nonparametric (BNP) methods (Ferguson, 1973; Müller and Mitra, 2013; Müller and Rodriguez, 2013) are gaining popularity in longitudinal data analysis and treatment effect modeling since they are characterized by parameters that live in an infinite-dimensional space, allowing one to flexibly approximate arbitrary distributions. For flexible longitudinal data analysis, Silva (2016) uses Gaussian process to model longitudinal outcome under different levels of interventions. In another example, Chib and Hamilton (2002) use the Dirichlet Process prior to add flexibility in representing the outcome and the treatment effects.

A number of related works have focused on heterogeneous treatment effects (HTE) by estimating the effects conditional on covariates defining subpopulations. For example, Tian et al. (2014) and Imai et al. (2013) apply regularized linear regression to select covariates characterizing subpopulations with differential outcomes, Johansson et al. (2016) use neural network to learn a multi-layer representation of high-dimensional covariates with differential outcomes. Other work use tree structure to partition based on covariates that identify subpopulations with different outcomes (Foster et al., 2010; Su et al., 2009) or different conditional treatment effects (Athey and Imbens, 2015). All of the above-mentioned works focus on obtaining point-in-time estimates. Only recently, Huang et al. (2015) and Xu and Ji (2014) have used parametric models to estimate treatment effects over time.

The proposed method advances state-of-the-art in a number of ways. First, in contrast with past works that focus on modeling response at a point-in-time, this work obtains the continuous response over time. Further, we obtain longitudinal responses from sparse and irregularly sampled observational data. Second, the proposed BNP model flexibly models variations in treatment effects while borrowing information across individuals. In applications such as education and healthcare where response across individuals can vary widely, recovering individual level effects is more informative for decision-making. Third, the fully Bayesian approach quantifies uncertainty at the individual level; this is particularly important for individualization where the estimated effects maybe uncertain due to lack of data. A key practical advantage of using nonparametric approaches is that they often provide better fits to challenging data than using parametric model based methods. This is particularly important in our application of estimating treatment response curves for physiologic time series.

---

1. This paper extends the method presented in (Xu et al., 2016).

## 2. Longitudinal Treatment Response Model

As a running example, we use the application of estimating the longitudinal outcome for creatinine level, an indicator of kidney function. Specifically, our goal is to obtain an individualized estimate of the response over time for treatments given for modulating creatinine level. We consider the problem of estimating the treatment responses from sparse, irregularly sampled data such as those in electronic health records (EHRs). There are two key challenges that must be addressed. First, in clinical data contained within EHRs, measurements are often not obtained at regular intervals, and measurement schedules vary across individuals. For example, caregivers may choose to make measurements once a day on some patients while multiple times a day on others. When the data are collected at fixed regular intervals, discrete-time approaches that maintain estimates only at specific points-in-time are adequate (e.g. Taubman et al. (2009)). To address this, we will employ functional representations instead (Quintana et al., 2015). Another key challenge is the presence of *time-varying confounding* (Robins, 1986, 1987). To correct for this confounding, our estimation is based on Robin's G-computation formula (Robins, 1986, 1987), a widely used approach in estimating treatment effects from sequential data with time-varying confounding.

**Time-varying confounding**: To understand time-varying confounding, let us first consider the simple example where a treatment tends to be assigned to sicker patients. Since these patients are sicker and also more likely to die, without accounting for the assignment bias, one might erroneously conclude this treatment is inferior. In the sequential-treatment assignment setting, such confounding occurs because doctors use the measurement of a variable to determine whether or not to treat, which in turn affects the variable's value at a subsequent time. The casual graph is presented on the left of Figure 1. In the graph, $Y$ denotes the final outcome, $L_0$, $L_1$, and $L_2$ denote the intermediate measurements or covariates, and $A_1$ and $A_2$ denote the treatments. From observational data, since we can only observe one treatment regime and one final outcome $Y$ for each patient, we apply Neyman-Rubin's causal model (Sekhon, 2008) to define potential outcomes for the unobserved counterfactuals. The model defines potential outcome $Y(a_1, a_2)$ as the outcome when treatment variables $A_1$ and $A_2$ are assigned to the values $a_1$ and $a_2$, respectively.

To adjust for time-varying confounding, G-computation formula makes two assumptions: consistency and conditional ignobility. First, the potential outcomes are assumed to be consistent with the observed outcomes, that is $Y(a_1, a_2) = (Y|A_1 = a_1, A_2 = a_2)$. Second, the treatment received at each time is randomly assigned (i.e. ignorable) conditional on past treatments and covariate history, that is $Y(a_1, a_2) \perp A_1, A_2 | L_0, L_1, L_2$. As a result, we can obtain a new causal graph on the right of Figure 1. Formally, we can write the conditional probability of potential outcome as

$$p(Y(a_1, a_2)|L_0, L_1, L_2) = p(Y(a_1, a_2)|A_1, A_2, L_0, L_1, L_2) \qquad (1)$$
$$= p(Y|A_1 = a_1, A_2 = a_2, L_0, L_1, L_2),$$

where the first equality comes from the conditional ignobility and the second equality comes from the consistency assumption. This is known as the likelihood component in the G-computation formula. Below we introduce the notations used in the rest of the paper, and propose the longitudinal treatment response model based on Eq. (1).

**Notation:** Assume we have observations $\boldsymbol{Y}_i = \{Y_{ij} : j = 1,..,J_i\}$ from the $i$th individual at (irregularly-sampled) times $\{t_{i1}, ..., t_{iJ_i}\}$. In addition, we have $\boldsymbol{X}_i = \{X_{ij} : j = 1,..,J_i\}$, where $X_{ij}$ is a $1 \times p$ vector of observed covariates (e.g., age, gender, observation times) about this individual. We also have treatments $\boldsymbol{A}_i = \{A_{il} : l = 1, ..., L_i\}$ that were given to patient $i$ at times $\{\tau_{i1}, ..., \tau_{iL_i}\}$, where $A_{il} = d$ for some treatment type $d \in \{1,..,D\}$. The value of a measurement $\boldsymbol{Y}_i$ within an interval $(t, T]$ is denoted by $\boldsymbol{Y}_{i,(t,T]}$. The sets of measurements and treatments preceding a time $t$ are denoted by $\boldsymbol{Y}_{i,<t}$ and $\boldsymbol{A}_{i,<t}$, respectively.

Our goal is to obtain posterior inference for the treatment response curves at the individual and population levels, and for the potential outcomes $\boldsymbol{Y}_{i,>t}$ given any sequence of treatments conditioned upon historical data. In contrast with prior methods that assume a parametric model for the potential outcome (e.g., Hernán et al. (2000)), we propose below a Bayesian nonparametric model that models longitudinal responses in three parts: a baseline progression with no treatments prescribed, responses to treatments over time, and noises. We tackle the general setting of learning from data with multiple exposures to the same treatment or different treatments under the assumption of additive treatment effects.

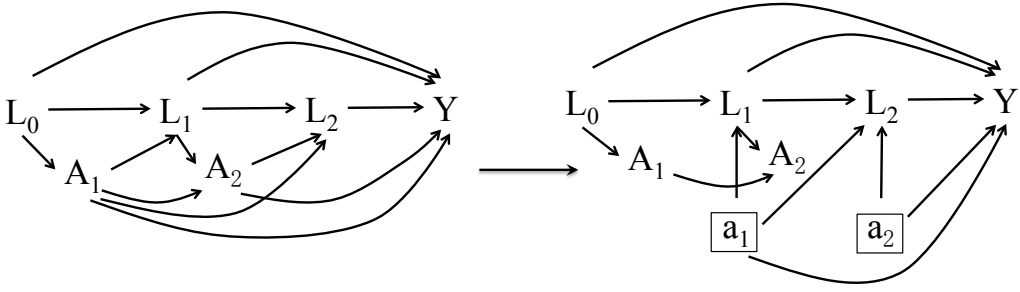

Figure 1: Causal graph for adjusting time-varying confounding by the G-computation formula.

We model the potential outcome $Y_{ij}$ using a generalized mixed-effects model combining the baseline progression and the treatment responses as follows:

$$Y_{ij}|X_{ij}, A_{i,<t_{ij}} = \underbrace{b(X_{ij}) + \boldsymbol{u}_i(t_{ij})}_{\text{baseline progression}} + \underbrace{\boldsymbol{f}_i(t_{ij}; A_{i,<t_{ij}})}_{\text{treatment response}} + \underbrace{\boldsymbol{\epsilon}_i(t_{ij}; A_{i,<t_{ij}})}_{\text{noise}}, \; j = 1,...,J_i. \qquad (2)$$

## 2.1 Modeling Baseline Progression

$b(X_{ij})$ is the fixed-effects component that captures the dependence of the outcome variable on the observed covariates $X_{ij}$. The features include time-invariant measurements (e.g., age, gender), denoted by $X_{i0}$, and time-varying measurements (e.g., observation times, changes in physiology), denoted by $X_{i1}(t_{ij})$. Here we model $b(X_{ij})$ as a linear regression:

$$b(X_{ij}; \boldsymbol{\beta}_i) = X_{ij}^T \boldsymbol{\beta}_i = X_{i0}^T \boldsymbol{\beta}_{i0} + X_{i1}(t_{ij})^T \boldsymbol{\beta}_{i1}. \qquad (3)$$

$\boldsymbol{u}_i(t_{ij})$ is the random-effects component that models the individual-specific deviations from $b(X_{ij})$ over time in baseline progression. We choose $\boldsymbol{u}_i$ to be generated from a zero-mean Gaussian process with a structured covariance $\mathcal{K}_{ui}(\sigma_{ui}^2, \rho_{ui}) = Cov(\boldsymbol{u}_i(t_{ij}), \boldsymbol{u}_i(t_{ij'})) = \sigma_{ui}^2 \rho_{ui}^{|t_{ij}-t_{ij'}|}$. Here, $\rho_{ui} \in (0,1)$. This represents an exponential covariance function, where $\sigma_{ui}^2$ is referred as a scale parameter and $\rho_{ui}$ as a smooth parameter. Similar choices were made by Quintana et al. (2015) in their application of modeling functional data. A different choice for both the mean and the covariance kernel can be made depending on the properties of the data; see Schulam and Saria (2015) for a different example of the baseline model for modeling progression in chronic diseases.

## 2.2 Modeling Treatment-Response

We focus on scenarios where treatment choices are discrete and treatment effects are additive. Given the set of treatments $A_{i,<t_{ij}}$ preceding time $t_{ij}$, we formulate the treatment response model as:

$$\boldsymbol{f}_i(t_{ij}; A_{i,<t_{ij}}) = \sum_{l:\tau_{il}<t_{ij}} g_{i,A_{il}}(t_{ij} - \tau_{il}), \qquad (4)$$

where $g_{i,A_{il}}(t_{ij} - \tau_{il})$ denotes the response curve of individual $i$ for treatment $A_{il}$ that was given at time $\tau_{il}$. To estimate the cumulative effect at $t_{ij}$, the response curves from the treatment set $A_{i,<t_{ij}}$ are added. We parameterize the function $g_{id}(t)$ as

$$g_{id}(t) = \begin{cases} b_0 + \alpha_{1_{id}}/[1 + \exp(-\alpha_{2_{id}}(t - \gamma_{id}/2))], & \text{if } 0 \leq t < \gamma_{id} \\ b_{id} + \alpha_0/[1 + \exp(\alpha_{3_{id}}(t - 3\gamma_{id}/2))], & \text{if } t \geq \gamma_{id}, \end{cases} \qquad (5)$$

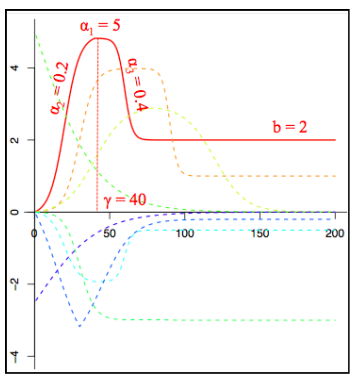

Figure 2: Examples of treatment response curves

with five free parameters $\{\alpha_1, \alpha_2, \alpha_3, \gamma, b\}$; here, the collection of individual-specific treatment response parameters $\alpha_{1_{id}}$'s are short-handed to $\alpha_1$ and so on.

The motivation for choosing this particular form of the $g_{id}(t)$ function is to obtain a flexible asymmetric "U" shape curve, as shown in Figure 2. We concatenate two sigmoid curves and allow the parameters for the two sigmoid functions and the point of switching between the two sigmoids to vary so that it can flexibly capture responses where a marker may either increase or decrease and eventually converges to a stable value. In Figure 2, we present several examples of such curves, and highlight one particularly for $g(t; \alpha_1 = 5, \alpha_2 = 0.2, \alpha_3 = 0.4, \gamma = 40, b = 2)$. Here, $\alpha_1 \in \Re$ represents the curve's maximum value and the sign of $\alpha_1$ determines whether the treatment increases (i.e. $\alpha_1 > 0$) or decreases (i.e. $\alpha_1 < 0$) the target marker value. $\alpha_2 \in (0,1)$ and $\alpha_3 \in (0,1)$ individually model the "steepness" of the two sigmoid curves; $\gamma \in \Re$ denotes the switching point; $b$ denotes the value that the curve stabilizes and is constrained such that $b/g(\gamma) \in (0,1)$. Lastly, to make the $g_{id}(t)$ function well defined, we set $b_0 = -\alpha_{1_{id}}/[1 + \exp(\alpha_{2_{id}}\gamma_{id}/2)]$ for attaining $g_{id}(0) = 0$, and $\alpha_0 = (a_{1_{id}} + 2b_0 - b_{id})/(1 + \exp(-a_{3_{id}}\gamma_{id}/2))$ for attaining a unique peek value at $t = \gamma_{id}$.

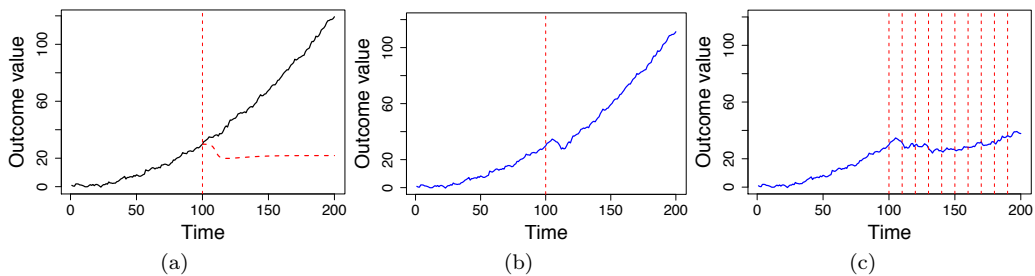

Figure 3: An illustration of additive treatment responses over time in single regime. (a) In black, an example baseline progression for the target outcome under no treatment. In red, we show the time of treatment assignment and the treatment-response curve. (b) In blue, the outcome over time is shown—obtained by adding the baseline progression and the treatment response curve. (c) The outcome over time when the treatment is assigned multiple times—the timing of the assignments are shown with vertical red lines.

In Figure 3, we illustrate the cumulative effects by adding multiple treatment responses. In Figure 3 (a), the black line denotes the increasing outcome due to an increasing baseline progression under no treatment. For example, in individuals with chronic kidney disease, their kidney function markers under no treatment become worse over time. In Figures 3 (a-c), the vertical red lines denote times when the treatment was prescribed. The reduced outcome values in the case of a single treatment versus multiple sequential treatments are shown in blue in Figures 3 (b) and 3 (c) respectively. Figures 3 (b-c) illustrate the cumulative effect from multiple treatment assignments.

### 2.3 Modeling Noise

We model the noise in two parts: the independent random noise $\epsilon_{ij}^0$ for individual $i$ at each time point $t_{ij}$ and the time-dependent random noise $\epsilon_d'(t)$ for treatment type $d$ within its effective window $W_d$. We model this additional noise because adding the effects of treatments can introduce higher uncertainty and error into the outcome model. The noise is time-dependent because the uncertainty reduces as the time from treatment administration increases and the effects diminish. So we formulate the overall noise as

$$\boldsymbol{\epsilon_i}(t_{ij}; \mathcal{H}_{ij}) = \epsilon_{ij}^0 + \sum_{l:\tau_{il}<t_{ij}} \epsilon'_{A_{il}}(t_{ij} - \tau_{il}), \tag{6}$$

where $\epsilon_{ij}^0$'s are i.i.d. Gaussian distributed with mean zero and variance $\sigma_{\epsilon_i}^2$, $\epsilon_d'(t)$'s are jointly Gaussian distributed with mean zero and structured covariance $\mathcal{K}_{\epsilon_d}(\sigma_{\epsilon_d'}^2, \rho_{\epsilon_d'}) = Cov(\epsilon_d'(t), \epsilon_d'(t')) = \sigma_{\epsilon_d'}^2 \rho_{\epsilon_d'}^{|t-t'|}$.

### 2.4 A Hierarchical Prior for Estimating Individualized Treatment Response (ITR) Curves

We posit a nonparametric hierarchical prior on the parameters in the baseline progression and the treatment response model. At the population level, this allows our model to flexibly capture heterogeneity in treatment response across individuals without specifying the number of subpopulations. Further, our model can borrow information across individuals to estimate one individual's treatment response by a hierarchical prior. It is worth noting that overfitting or lack of reliable estimates is a key concern when fitting treatment responses at the individual level. Within the proposed approach, we obtain the full posterior distribution over the individual's treatment response rather than a point estimate for the parameters. Further, when little data are available on an individual, their ITR estimate is similar to that of the population. However, as more data are collected, the posterior distribution over the ITR becomes more informative regarding if and how the individual's ITR differs from other individuals.

Bayesian nonparametric approaches such as Dirichlet process (DP) and DP mixture have been widely used in clustering time-series data. For example, Ren et al. (2015) have applied DP priors in latent factor models to cluster multiple housing price data streams. Nieto-Barajas et al. (2014) uses a generalization of DP mixture—Possion-DP priors—in linear dynamic models to group stock exchange data. We use the DP mixture to cluster the parameters for the fixed-effects component of the baseline progression and the treatment response curves, and use the DP mixture of Gaussian processes to cluster the random-effects component for the baseline progression. The DP mixture of GPs has been used to identify and group the sub-divisions in each individual's gene expression (Hensman et al., 2015) or disease trajectories (Ross and Dy, 2013). We use DP mixture of GPs to group individuals' baseline deviations on their similarity in the GP's kernel parameters.

#### 2.4.1 BACKGROUND ON THE DIRICHLET PROCESS MIXTURE

We briefly describe the DP and the DP mixture (DPM). Ferguson (1973) introduced the DP prior as a probability distribution on an infinite dimensional measurable space of probability measures. The stick-breaking construction by Sethuraman (1994) provides an intuitive and interpretable representation of the DP. Let $G_0$ be a known distribution and let $M > 0$ be a positive constant.

Then we say $G \sim DP(G_0, M)$ provided

$$G(\cdot) = \sum_{k=1}^{\infty} \omega_k \delta_{\theta_k}(\cdot), \ \theta_k \overset{iid}{\sim} G_0,$$

where $\delta_{\theta_k}(\cdot)$ defines point mass at $\theta_k$ and $\omega_k$'s are defined as

$$\omega_k = V_k \prod_{r=1}^{k-1}(1 - V_r), \ V_k \sim Beta(1, M).$$

Thus $G$ is a random distribution that is discrete with probability one. $G_0$ is the base or centering distribution since $E(G) = G_0$. The discrete nature of the DP makes it inappropriate for modeling continuous data where units within a partition share similar rather than the same parameter. Therefore, DPM extends DP by introducing a continuous kernel centered at $\theta_k$ instead of a point mass $\delta_{\theta_k}$. Let $y_1, y_2, \ldots$ be i.i.d. samples and $f(\cdot|\theta)$ be a parametric density function, we can write the stick-breaking construction of the DPM as

$$y_i \mid (\omega_k), (\theta_k) \sim \sum_{k=1}^{\infty} \omega_k f(\cdot|\theta_k), \ \theta_k \sim G_0.$$

2.4.2 HIERARCHICAL INDIVIDUALIZED TREATMENT-RESPONSE (ITR) MODEL

We leverage the DPM prior to cluster both the baseline progression and the treatment response parameters—while allowing individual-specific variability—and obtain a hierarchical treatment-response model as shown in Figure 4. Specifically, let $\boldsymbol{b}_i$ denote the sum of the fixed-effects component $b(\boldsymbol{X}_i^M)$ and the random-effects component $\boldsymbol{u}_i$ in the baseline progression. Then based on the description in Section 2.1, $\boldsymbol{b}_i$ follows the distribution

$$p(\boldsymbol{b}_i|\boldsymbol{\varphi}_i) = \mathcal{N}((\boldsymbol{X}_i^M)^T \boldsymbol{\beta}_i, \mathcal{K}_{ui}), \tag{7}$$
$$\mathcal{K}_{ui}(t_{ij}, t_{ij'}; \sigma_{ui}^2, \rho_{ui}) = \sigma_{ui}^2 \rho_{ui}^{|t_{ij} - t_{ij'}|},$$

where $\boldsymbol{\varphi}_i = \{\boldsymbol{\beta}_i, \sigma_{ui}^2, \rho_{ui}\}$ denotes all the individual-specific baseline progression parameters. We posit a DPM prior on $\boldsymbol{\varphi}_i$'s, and obtain the following distributions

$$p(\boldsymbol{\varphi}_i) = \sum_{k=1}^{\infty} \omega_k \mathcal{N}(\boldsymbol{\beta}_i; \boldsymbol{\beta}_{b_k}^*, \Sigma_{b_k}^*) \frac{\mathcal{N}(\log(\sigma_{ui}^2); \mu_{\sigma'_{uk}}^*, \sigma_{\sigma'_{u0}}^2)}{\sigma_{ui}^2} \frac{\mathcal{N}(\text{logit}(\rho_{ui}); \mu_{\rho'_{uk}}^*, \sigma_{\rho'_{u0}}^2)}{(1-\rho_{ui})^2},$$
(8)

$$\omega_k = V_{\varphi_k} \prod_{r=1}^{k-1}(1 - V_{\varphi_r}),$$
$$p(V_{\varphi_k}) = \text{Beta}(1, M_1),$$
$$p(\boldsymbol{\beta}_{b_k}^*, \Sigma_{b_k}^*, \mu_{\sigma'_{uk}}^*, \mu_{\rho'_{uk}}^*) = \text{NIW}(\boldsymbol{\beta}_{b_k}^*, \Sigma_{b_k}^*; \beta_0, \kappa_0, \theta_0, \boldsymbol{S}_0) \mathcal{N}(\mu_{\sigma'_{uk}}^*; \mu_{\sigma'_0}, \sigma_{\sigma'_0}^2) \mathcal{N}(\mu_{\rho'_{uk}}^*; \mu_{\rho'_0}, \sigma_{\rho'_0}^2).$$

For parameters that lie in real space we assume they are sampled from a Gaussian distribution. For parameters that are constrained, such as $\sigma_{ui}^2 \in (0, +\infty)$ and $\rho_{ui} \in (0, 1)$, we transform the support of these variables into real space first and posit Gaussian priors on the transformed unconstrained variables. This requires a calculation of the additional Jacobian adjustment $|\det J(T^{-1}(y))|$ for each transformation $y = T(x)$ (Olive, 2014). A detailed description of deriving the Jacobian adjustment is given in Appendix A.

We let $\boldsymbol{\theta}_{\varphi_k}^* = \{\boldsymbol{\beta}_{b_k}^*, \Sigma_{b_k}^*, \mu_{\sigma'_{uk}}^*, \mu_{rho'_{uk}}^*\}$ denote the component-specific parameters that the transformed $\boldsymbol{\varphi}_i$'s are centered at. We introduce a discrete latent variable $Z_{\varphi_i}$ to indicate the mixture component associated with individual $i$, and sample the $Z_{\varphi_i}$'s from a Multinomial distribution over the weight $\omega_k$'s. The hyperparameter $M_1$ controls the degree of clustering and generates the $V_{\varphi_k}$'s to formulate the associated component weight $\omega_k$'s. Lastly, we let $\boldsymbol{a}_1 = \{\beta_0, \kappa_0, \theta_0, \boldsymbol{S}_0, \mu_{\rho'_0}, \sigma^2_{\rho'_0}, \mu_{\sigma'_0}, \sigma^2_{\sigma'_0}\}$ denote all the hyperparameters that are used in $\boldsymbol{\theta}^*_{\varphi_k}$'s base distribution. In our applications, as discussed later in Section 4.2 and 4.3, these hyperparameters are chosen to construct broad (and uninformative) base distributions.

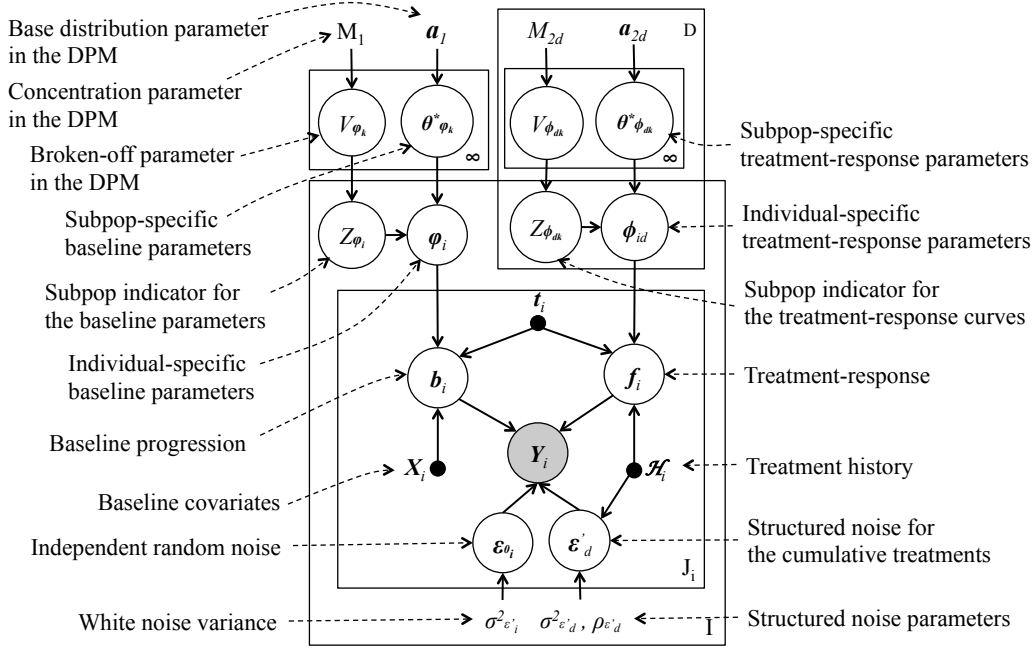

Figure 4: Graphical representation of the hierarchical treatment response model. Hidden variables are circled; observed outcome variables are shaded; observed input variables are filled.

For the treatment response model, $\boldsymbol{f}_i$ is defined as Eq. (4). We let $\boldsymbol{\phi}_{id} = \{\alpha_{1_{id}}, \alpha_{2_{id}}, \alpha_{3_{id}}, \gamma_{id}, b_{id}\}$ denote the individual-specific treatment parameters in the $g_{id}(t)$ function in Eq. (5). Similarly, we put DPM priors on $\boldsymbol{\phi}_{id}$'s, and obtain the following distributions

$$p(\boldsymbol{\phi}_{id}) = \sum_{k=1}^{\infty} \omega_{dk} \mathcal{N}(T^{-1}(\boldsymbol{\phi}_{id}); \boldsymbol{\mu}^*_{\phi'_{dk}}, \boldsymbol{D}_{\phi'_0}) |\zeta_{id}|/(1-\alpha_{2_{id}})^4, \qquad (9)$$

$$\omega_{dk} = V_{\phi_{dk}} \prod_{r=1}^{k-1} (1 - V_{\phi_{dk}}),$$

$$p(V_{\phi_{dk}}) = \text{Beta}(1, M_{2d}),$$

$$p(\boldsymbol{\mu}_{\phi'_{id}}) = \mathcal{N}(\boldsymbol{\mu}_{\phi'_{id}}; \boldsymbol{\mu}_{d_0}, \boldsymbol{D}_{d_0}),$$

where $T^{-1}(\boldsymbol{\phi}_{id}) = \{\alpha_{1_{id}}, \text{logit}(\alpha_{2_{id}}), \text{logit}(\alpha_{3_{id}}), \gamma_{id}, \text{logit}(b_{id}/g(\gamma_{id}))\}$ is a real-space vector transformed from $\boldsymbol{\phi}_{id}$'s, and $\zeta_{id} = -1/g(\gamma_{id})(1 - b/g(\gamma_{id}))^2$ is the Jacobian adjustment computed from the transformation. A detailed description is given in Appendix A.

We let $\boldsymbol{\theta}^*_{\phi_{id}} = \{\boldsymbol{\mu}^*_{\phi'_{dk}}\}$ denote the component-specific parameters that $T^{-1}(\boldsymbol{\phi}_{id})$'s are centered at. We introduce a discrete latent variable $Z_{\phi_{id}}$ to indicate the mixture component associated with individual $i$'s response to dose type $d$, and sample $Z_{\phi_{id}}$'s from a Multinomial distribution over the weight $\omega_{dk}$'s. The hyperparameter $M_{2d}$ controls the degree of clustering and generates the $V_{\phi_k}$'s to formulate the associated component weight $\omega_{dk}$'s. We let $\boldsymbol{a}_{2d} = \{\boldsymbol{\mu}_{d_0}, \boldsymbol{D}_{d_0}\}$ denote all the hyperparameters that are used in $\boldsymbol{\theta}^*_{\phi_{id}}$'s base distribution. In our applications, as discussed later in Section 4.2 and 4.3, these hyperparameters are chosen to construct broad (and uninformative) base distributions.

For the noise model, based on the description in Section 2.3, $\boldsymbol{\epsilon}_i$ follows the distribution

$$p(\boldsymbol{\epsilon}_i | \sigma^2_{\epsilon_i}, \boldsymbol{\sigma}^2_{\epsilon'}, \boldsymbol{\rho}_{\epsilon'}) = \mathcal{N}(\boldsymbol{0}, \sigma^2_{\epsilon_i} \boldsymbol{I}_{J_i} + \mathcal{K}_{\epsilon'_i}), \tag{10}$$

$$\mathcal{K}_{\epsilon'_i}(t_{ij}, t_{ij'}; \boldsymbol{\sigma}^2_{\epsilon'}, \boldsymbol{\rho}_{\epsilon'}) = \begin{cases} \sum_l \sigma^2_{\epsilon'_{A_{il}}} \rho_{\epsilon'_{A_{il}}}^{|t_{ij}-t_{ij'}|} & , \forall l \text{ s.t. } t_{ij} > \tau_{il}, t_{ij'} > \tau_{il}, \\ 0 & , \text{ otherwise.} \end{cases}$$

To complete the prior specification, we put inverse gamma (IG) on the scale parameter $\sigma^2_{\epsilon_i}$, and transform the constrained parameters $\sigma^2_{\epsilon'_d}$ and $\rho_{\epsilon'_d}$ into real space then posit Gaussian priors on the transformed unconstrained variables:

$$\begin{aligned}p(\sigma^2_{\epsilon_i}) &= \text{IG}(s_\epsilon, \nu), \\ p(\sigma^2_{\epsilon'_d}) &= \mathcal{N}(\log(\sigma^2_{\epsilon'_d}); \mu_{\epsilon_1}, \sigma^2_{\epsilon_1})/\sigma^2_{\epsilon'_d}, \\ p(\rho_{\epsilon'_d}) &= \mathcal{N}(\text{logit}(\rho_{\epsilon'_d}); \mu_{\epsilon_2}, \sigma^2_{\epsilon_2})/(1-\rho_{\epsilon'_d})^2.\end{aligned} \tag{11}$$

## 3. Inference

We use MCMC to approximate the posterior distributions of the parameters in the proposed model. Consider the joint posterior

$$\begin{aligned}&p(\boldsymbol{u}, \boldsymbol{f}, \boldsymbol{\varphi}, \boldsymbol{\phi}, \sigma^2_\epsilon, \boldsymbol{\sigma}^2_{\epsilon'}, \boldsymbol{\rho}_{\epsilon'} \mid \boldsymbol{Y}) \\ &\propto \left\{ \prod_{i=1}^I p(\boldsymbol{Y}_i, \boldsymbol{b}_i, \boldsymbol{f}_i \mid \boldsymbol{\varphi}_i, \boldsymbol{\phi}_i, \sigma^2_{\epsilon_i}, \boldsymbol{\sigma}^2_{\epsilon'}, \boldsymbol{\rho}_{\epsilon'}) \right\} p(\boldsymbol{\varphi}) p(\boldsymbol{\phi}) p(\boldsymbol{\sigma}^2_{\epsilon_i}) p(\boldsymbol{\sigma}^2_{\epsilon'}) p(\boldsymbol{\rho}_{\epsilon'}) \\ &= \left\{ \prod_{i=1}^I p(\boldsymbol{Y}_i \mid \boldsymbol{b}_i, \boldsymbol{f}_i, \sigma^2_{\epsilon_i}, \boldsymbol{\sigma}^2_{\epsilon'}, \boldsymbol{\rho}_{\epsilon'}) p(\boldsymbol{b}_i \mid \boldsymbol{\varphi}_i) p(\boldsymbol{f}_i \mid \boldsymbol{\phi}_i) \right\} p(\boldsymbol{\varphi}) p(\boldsymbol{\phi}) p(\boldsymbol{\sigma}^2_\epsilon) p(\boldsymbol{\sigma}^2_{\epsilon'}) p(\boldsymbol{\rho}_{\epsilon'}),\end{aligned} \tag{12}$$

the first term in the product is $\mathcal{N}(\boldsymbol{X}_i \boldsymbol{b}_i + \boldsymbol{f}_i, \sigma^2_{\epsilon_i} \boldsymbol{I}_{J_i} + \mathcal{K}_{\epsilon'_i})$ with $\mathcal{K}_{\epsilon'_i}$ specified in Eq. (10), the second term is defined in Eq. (7), the third term is deterministic and specified in Eq. (4). We factorize the remaining terms as

$$\left\{ \prod_{i=1}^I \{p(\boldsymbol{\varphi}_i) \prod_{d=1}^D p(\boldsymbol{\phi}_{id})\} \right\} \left\{ \prod_{i=1}^I p(\sigma^2_{\epsilon_i}) \right\} \left\{ \prod_{d=1}^D p(\sigma^2_{\epsilon'_d}) p(\rho_{\epsilon'_d}) \right\},$$

with each distribution specified in Eq. (8, 9, and 11) respectively.

For the infinite-dimensional DPM priors on $\boldsymbol{\varphi}_i$ and $\boldsymbol{\phi}_{id}$, we approximate these using a truncated stick-breaking process that was developed by Ishwaran and James (2001). Ishwaran and James (2001) justify that the truncated process greatly reduces computations and can closely approximate a full Dirichlet process when the truncation level is large relatively to the number of observations. Based on Theorem 2 in Ishwaran and James (2001), for experiments where we have fewer than 500 time series, a truncation level of 20 for the number of clusters provides an approximation with error bound of $1.12e-05$ on the $L_1$ distance between the true marginal distribution and the approximate marginal distribution of the data. Given the truncation approximation, it allows us to use standard MCMC algorithms to update the parameters in the finite-dimension space. Particularly, we develop a Gibbs sampler. We posit conjugate priors on the parameters; this allows us to obtain the conditional distributions for the variables within each Gibbs loop in closed form. Specifically, the forms of these conditional distributions for the component indicator variables $Z_{\varphi_i}$ and $Z_{\phi_{id}}$ are Multinomial, for the component-level variables $\beta^*_{b_k}$, $\Sigma^*_{b_k}$ are Normal-inverse-Wishart, for $\mu^*_{\sigma'_{u_k}}$, $\mu^*_{\rho'_{u_k}}$ and $\mu^*_{\phi_{d_k}}$ are Gaussian, for $V_{\varphi_k}$ and $V_{\phi_{dk}}$ are Beta, and for the concentration parameters $M_1$ and $M_{2d}$ are Gamma. In addition to the DPM related parameters, we also use Gibbs sampler to infer $\boldsymbol{\beta}_i$, $\boldsymbol{b}_i$ and $\sigma^2_{\epsilon_i}$. The forms of these conditional distributions are all Gaussian. Detailed descriptions of deriving the Gibbs samplers are in Appendix B.1 and B.2.

For the remaining variables, the conditional distributions cannot be derived in closed form. Thus we use a Metropolis-Hastings sampler. Particularly, for the unconstrained variables $\alpha_{1_{id}}$ and $\gamma_{id}$, we choose the proposal distribution to be normal with standard deviation of 0.3. For the constrained variables $\sigma^2_{ui} \in (0, +\infty)$ and $\sigma^2_{\epsilon'_d} \in (0, +\infty)$, we choose the *truncated normal* distribution (Greene, 2003) with standard deviation of 0.3. For variables $\rho_{ui}$, $\alpha_{2_{id}}, \alpha_{3_{id}}$ and $b_{id}/g(\gamma_{id})$ that are constrained in $(0, 1)$, we choose normal with standard deviation 0.15 to propose a new sample. Then, we reflect the new sample by 0 or 1, probably multiple times if needed, to map it back onto the support $(0, 1)$. A detailed description of the Metropolis-Hasting sampler is in Appendix B.3. In our experiments, which will be discussed in Section 4.2, these proposal distributions performed well (with acceptance rates ranging from 0.14 - 0.22 across the chains).

## 4. Numerical Results

We evaluate the proposed model in three experiments, including a simulation study and two observational studies. We first conduct a simulation study to evaluate whether the proposed model can uncover the true treatment response curves and the extent to which the quality of the recovered curves depends on the choice of the hyperparameters. Then, we evaluate the model's performance on estimating treatment responses using two observational data sets. With the first dataset, we estimate the individual's response curves to treatments for managing creatinine levels. High creatinine levels can indicate kidney function deterioration, and renal replacement therapy (RRT) can be given for treatment. With the second dataset, we estimate the effects of diuretics on fluid balance. In conditions of critical illness, the body's ability to remove excess water can deteriorate. Diuretics can be administered to remove the excess sodium and water from the body.

### 4.1 Simulation Study

**Datasets.** We simulate 200 trajectories with each trajectory's duration uniformly sampled from 18-24 hours. To sample observation times within a trajectory, we sample the time for the next observation uniformly from 5-15 minutes at each a given time. The resulting trajectories have on average 126 measurements. To sample treatment times, we sample the time for the next treatment uniformly from 60-80 minutes at each given time. At each time, we also sample the treatment assignment conditioned on previous assignments and outcomes. This satisfies the ignobility assumption described in Section 2. We generate three types of treatments including the following: no treatment, treatment 1 that has an increasing effect on the outcome, and treatment 2 that has a decreasing effect. We designate the conditional probabilities to maintain the outcome within a "normal" range with high probabilities. For example, given treatment 1 was prescribed at the previous time point, we assign high probability to generate treatment 1 again at the current time point if the outcome is still low, but assign low probability to generate treatment 1 if the outcome has been increased into the normal range. The detailed conditional probabilities are specified in Table 1, where $A_t$ denotes the treatment assignment at time $t$ (0 for no treatment, 1 for treatment 1, and 2 for treatment 2), and $Y_t$ indicates the outcome level at time $t$ (0 for below, 1 for within, and 2 for above the normal range). We define normal range as $[15, 25]$ in our simulation. Figure 5 shows one example trajectory: the black points are the observations of the outcome, the colored vertical dotted lines denote when the treatments are prescribed, and different colors refer to different treatment types. In average, we generate 9 treatments per trajectory.

| $A_t$ $A_{t-1}$ $Y_{t-1}$ | $P(A_t\|A_{t-1},Y_{t-1})$ | $A_t$ $A_{t-1}$ $Y_{t-1}$ | $P(A_t\|A_{t-1},Y_{t-1})$ | $A_t$ $A_{t-1}$ $Y_{t-1}$ | $P(A_t\|A_{t-1},Y_{t-1})$ |
|---|---|---|---|---|---|
| 0 0 0 | 0.3 | 0 1 0 | 0.3 | 0 2 0 | 0.6 |
| 1 0 0 | 0.5 | 1 1 0 | 0.6 | 1 2 0 | 0.3 |
| 2 0 0 | 0.2 | 2 1 0 | 0.1 | 2 2 0 | 0.1 |
| 0 0 1 | 0.8 | 0 1 1 | 0.8 | 0 2 1 | 0.8 |
| 1 0 1 | 0.1 | 1 1 1 | 0.1 | 1 2 1 | 0.1 |
| 2 0 1 | 0.1 | 2 1 1 | 0.1 | 2 2 1 | 0.1 |
| 0 0 2 | 0.3 | 0 1 2 | 0.6 | 0 2 2 | 0.1 |
| 1 0 2 | 0.2 | 1 1 2 | 0.1 | 1 2 2 | 0.1 |
| 2 0 2 | 0.5 | 2 1 2 | 0.3 | 2 2 2 | 0.8 |

Table 1: Conditional probability table for simulating treatment assignments

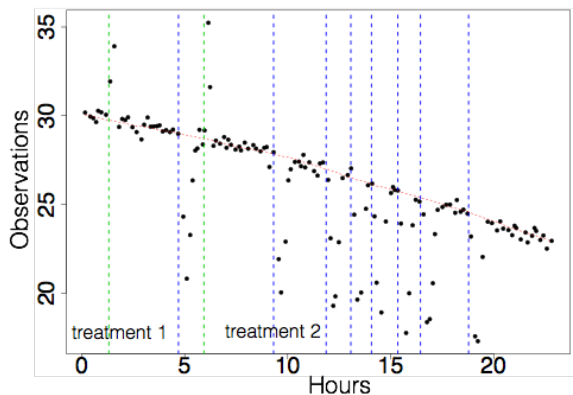

Figure 5: A simulated trajectory: the red dotted lines denote the simulated baseline progression, the vertical dotted lines denote the simulated treatments (green is treatment 1 and blue is treatment 2), and the black dots denote the resulting observations.

To sample observation values, we individually sample the baseline progression, the treatment responses, and the noise terms that are specified in Eq. 2. In the baseline progression, we choose feature $\boldsymbol{X}_{ij} = \{1, t_{ij}, t_{ij}^2\}$ for individual $i$ at observation time $t_{ij}$. We assume the parameters $\boldsymbol{\beta}_i$, $\sigma_{ui}^2$, and $\rho_{ui}$ are sampled from a three-mixture model. Specifically, the three-mixture component parameters for $\boldsymbol{\beta}_i$'s are $\{5, 5, 3\}$, $\{30, -5, -3\}$ and $\{10, -2, -1\}$, the component parameters for $\sigma_{ui}^2$'s are $0.1^2$, $0.1^2$, and $0.1^2$, and the component parameters for $\rho_{ui}$'s are 0.1, 0.9, and 0.5. We assign each individual to one of the three components with equal probabilities. Given the assignment to a component, we sample the individual's parameters from a Gaussian distribution centered at the component parameters with a small variance of $0.1^2$ ($\sigma_{ui}^2$ and $\rho_{ui}$ were transformed into real-space). In the left panel of Figure 6 (a), we show the resulting baseline progressions from the three-mixture components.

For each treatment, the treatment response parameters $\{\alpha_{1_{id}}, \alpha_{2_{id}}, \alpha_{3_{id}}, \gamma_{id}, b_{id}\}$ are sampled from a three-mixture model. The component parameters for treatment 1 are specified as $\{10, 0.9, 0.4, 10, 0\}$, $\{5, 0.9, 0.9, 5, 0\}$, and $\{8, 0.7, 0.7, 15, 0.001\}$; the component parameters for treatment 2 are specified as $\{-10, 0.9, 0.7, 20, 0\}$, $\{-6, 0.5, 0.5, 15, 0\}$, and $\{-8, 0.4, 0.3, 25, 0\}$. For each treatment, we assign each individual to one of the three components with equal probabilities. Given the assignment to a component, we sample the individual's treatment response parameters from a Gaussian distribution centered at the component parameters with a small variance of $0.3^2$ ($\alpha_{2_{id}}$, $\alpha_{3_{id}}$, and $b_{id}/g(\gamma_{id})$ were transformed into real-space). In the right two panels of Figure 6 (a), we show the resulting treatment response curves sampled from each of the three-mixture models.

For noise, we sample the i.i.d. noise from zero-mean Gaussian distribution with variance of $0.3^2$. For each given treatment, we sample the time-dependent noise from the zero-mean multivariate Gaussian distribution with the exponential kernel specified by $\sigma^2_{\epsilon'_d} = 0.1^2$ and $\rho_{\epsilon'_d} = 0.9$.

**Experimental setup.** For the fixed-effect component in the baseline progression, we posit a non-informative Normal-inverse-Wishart base distribution NIW($\mathbf{0}$, 1, p+2, $I_{p+2}$) on the regression coefficients, where $p = 3$ since we use 3 covariates in simulation. For the random-effect component in the baseline progression, we posit a non-informative Gaussian base distribution $\mathcal{N}(\text{logit}(0.5), 4)$ (covers range $(0.02, 0.98)$ with 95% confidence) on the transformed smooth parameters $\rho_{ui}$'s defined in the exponential Gaussian process kernel. GP estimation is known to be consistent in estimating latent functions with a known convergence rate (van der Vaart and van Zanten, 2009), but is asymptotically unidentifiable in estimating the parameters in the Matérn class covariance functions (including exponential covariance) with infill domain (Zhang, 2004). Thus, we conduct a sensitivity analysis to study the quality of the recovered curves based on different choices of these scale parameters. For real applications, as discussed in Section 4.2 and 4.3, we will posit strong priors on these parameters to make them concentrated at small values.

For the treatment response model, we posit a multivariate Gaussian base distribution on the transformed response parameters for each treatment type: $\mathcal{N}\big(\{8, \text{logit}(0.5), \text{logit}(0.5), 10, \text{logit}(0.5)\}, 4\mathbf{I}_5\big)$ for treatment 1 and $\mathcal{N}\big(\{-8, \text{logit}(0.5), \text{logit}(0.5), 20, \text{logit}(0.5)\}, 4\mathbf{I}_5\big)$ for treatment 2. We choose these base distributions to be non-informative except for the peak effects and the change points that are chosen based on prior knowledge. For real applications, as discussed in Section 4.2 and 4.3, we also choose these priors based on expert domain knowledge.

For the i.i.d noises, we posit a non-informative Inverse-Gamma prior IG(1,1) on their variances. For the time-dependent noises in treatment responses, we posit a non-informative prior $\mathcal{N}(\text{logit}(0.5), 4)$ on the transformed smooth parameters, and posit a strong prior $\mathcal{N}(\log(0.1^2), 0.3^2)$ on the transformed scale parameters to avoid identifiable problems.

We randomly initialize one chain and run it for $5,000$ iterations with a burn-in of $2,500$ iterations. For each individual, we estimate the baseline progression and the treatment response curves based on the parameters sampled at each iteration after the burn-in. Afterwards, we average over the $2,500$ estimates to obtain the mean baseline progression and the mean treatment response curves for each individual. To evaluate the model performance, we recover each individual's component assignments from the last iteration and compare the resulting distribution with the simulation truth.

**Results.** In Figure 6 (b-f) we present the recovered baseline progressions and treatment response curves based on different scale parameters. Both the individual estimates and the three-mixture component estimates in the baseline progression and the treatment response curves are accurately recovered when the scale parameter is set to be that of the truth (i.e. $0.1^2$). The individual baseline progression and treatment response curves are mostly recovered and the three-mixture component are still accurately recovered when the scale parameters are set to be close to the truth (i.e. $0.05^2$ or $0.5^2$). When the scale parameters are set to be large or unclamped, the model become too flexible so that the variations in the treatment responses could partially be explained by the variations of the baseline progressions themselves. Therefore, both the peak values and the change points in the treatment response curves cannot be estimated accurately.

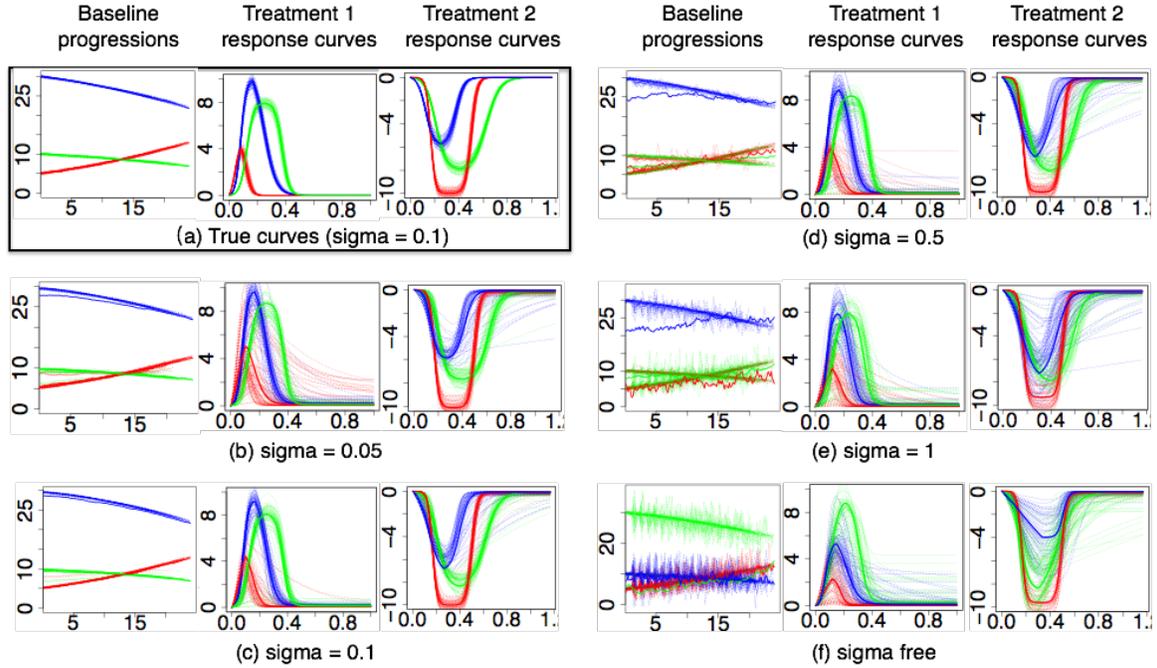

Figure 6: The true baseline progressions and treatment response curves used in simulation and the results recovered by the proposed model (different colors denote the different mixture components)

## 4.2 Observational Study: Estimating Heterogeneous Response Curves to Renal Replacement Therapy

We evaluate the proposed model on the task of estimating patients' responses to renal replacement therapy (RRT), a type of treatment to manage kidney function. Acute kidney injury (AKI) is a common complication in the intensive care unit (ICU). Furthermore, studies show that the mortality in AKI patients who depend on RRT can be as high as 80% (Tolwani, 2012; Bellomo et al., 1999). The decision to initiate RRT is complicated and not all the patients respond to RRT in the same manner. Thus, an individualized assessment tool is needed to determine which patients will benefit from RRT and which will not (Mehta, 2016).

**Datasets.** We fit our models on electronic health record data from patients admitted to the Beth Israel Deaconess Medical Center in Boston. The data are publicly available in the MIMIC (Multiparameter Intelligent Monitoring in Intensive Care)-II Clinical Database (Saeed et al., 2002). We estimate the effects of three types of RRT: Intermittent Hemodialysis (IHD), Continuous Veno-Venous Hemofiltration (CVVH), and Continuous Veno-Venous Hemodialysis (CVVHD). We use creatinine level as the patient's outcome. The creatinine level can increase when the kidney function declines. Figure 7 presents a typical trajectory from MIMIC-II patients who are prescribed RRTs. To decrease the creatinine levels, RRT can be initiated as an artificial replacement of the kidneys' function. After creatinine levels are decreased into the target range, RRT is discontinued and can be re-initiated when needed. Because RRT does not cure kidney disease permanently, creatinine levels rise again after a period of discontinuation on RRT. This is handled using a time-varying covariate as discussed in the experimental setup.

To select individuals from the database, we include the patients who were prescribed IHD, CVVH, and CVVHD during their ICU stay. We exclude the patients who have less than 15 observations of creatinine levels following the first elevated creatinine blood test measurement (i.e. creatinine level higher than 1.3 mg/dL for men and 1.1 mg/dL for women). The dataset contains 428 trajectories with a total of 16, 593 creatinine observations. Each individual trajectory has an average duration of 23 days. IHD are usually prescribed 3 times a week, each treatment lasting 3 to 6 hours or less (Pannu and Gibney, 2005). CVVH and CVVHD are two modalities of Continuous Renal Replacement Therapy (CRRT), and they are intended to be applied for 24 hours per day in ICU (Pannu and Gibney, 2005). The dataset contains a total of 525 instances of IHD, 186 of CVVH, and 981 of CVVHD. Creatinine levels were standardized by the population mean of 3.16 and standard deviation of 1.87.

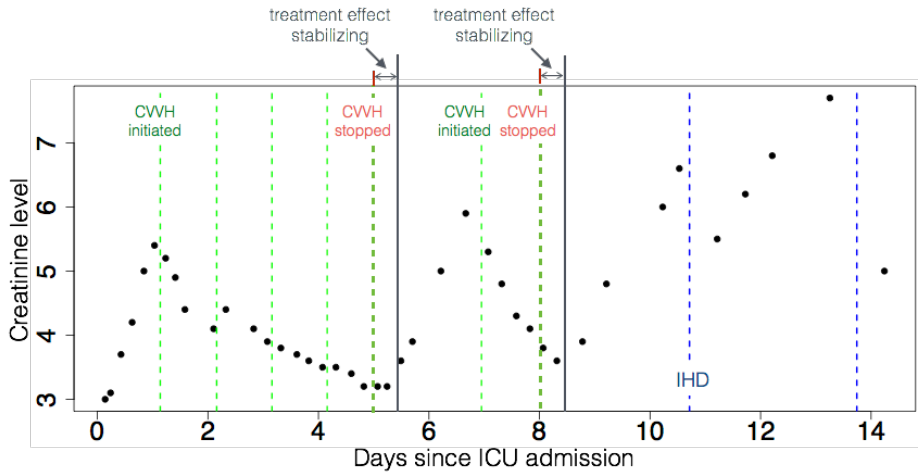

Figure 7: An example trajectory of managing creatinine levels within multiple treatment regimes. Black dots denote the observed creatinine levels. Vertical dashed lines denote the timing of treatments.

**Baselines.** We compare the performance of ITR with three baselines: the *pop* model, *individual* model, and *sub-pop* model. First, we evaluate against what we refer to as the *pop* model, which estimates treatment responses at the population level and does not take into account variations across individuals. The *pop* model is an instance of ITR where the baseline progression and the treatment response (transformed) parameters are drawn uniformly from a broad prior. To evaluate the extent to which individualizing the treatment response estimates is important, we also compare ITR against a second baseline called the *indivdiual* model. In the *individual* model, the parameters are drawn independently from a broad prior so that each individual samples its own set of parameters. Lastly, we compare ITR against a third baseline, called the *sub-pop* model, where the parameters are drawn from a DP instead of a DPM. This allows treatment responses to vary by subgroups but there is no explicit representation for differences across individuals within a subgroup.

**Experimental setup.** We assume that the fixed-effects component in the baseline progression is a linear regression model. We include the patient's age and weight as two baseline covariates. In addition, we include a time-varying covariate as follows. As described above, creatinine increases over time once RRT has been discontinued. This drift is modeled using a function of time—in this case, $\log(t - W)$, where $t$ is the time since last RRT discontinuation. $W$ is the window of time it takes for creatinine to stabilize after RRT discontinuation. $W$ was selected based on clinical guidance: $W = 6$ hours for IHD and $W = 12$ hours for CRRT. In Figure 7, we show an example creatinine trajectory and the example window W. Thus in total, we have $p = 4$ covariates (i.e., age, weight, time, and 1 for the intercept). We posit a non-informative Normal-inverse-Wishart base distribution $\text{NIW}(\mathbf{0}, 1, p + 2, I_p)$ for the regression coefficients. We posit a strong prior $\mathcal{N}(\log(0.1^2), 0.3^2)$ on the transformed scale parameters in the exponential Gaussian process kernel and a non-informative prior $\mathcal{N}(\text{logit}(0.5), 4)$ on the transformed smooth parameters.

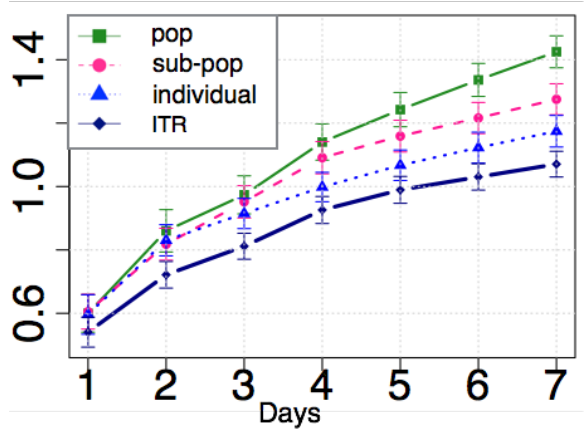

Figure 8: Comparison of ITR vs. baseline models prediction errors on creatinine.

For the treatment response model, we posit a Gaussian prior on the peak effects with the means chosen by the domain expert. Specifically, these were set as $\mathcal{N}(-2, 1)$ for IHD and $\mathcal{N}(-1, 1)$ for both CVVH and CVVHD. Similarly, we posit a Gaussian prior on the change points with the means chosen by the domain expert based on the expected duration over which the treatment takes effect. Specifically, these were set as $\mathcal{N}(1 \text{ hr}, 100 \text{ hrs})$ for IHD and $\mathcal{N}(12 \text{ hrs}, 100 \text{ hrs})$ for both CVVH and CVVHD. We posit non-informative Gaussian priors $\mathcal{N}(\text{logit}(0.5), 4)$ for the two steepness parameters and the ratio of the long-term effect to the peak effect.

For the noise, we posit an Inverse-Gamma prior $IG(1, 1)$ on the variances of the i.i.d. noises. We posit strong priors $\mathcal{N}(\log(0.1^2), 0.3^2)$ on the scale parameters for the time-dependent noises, and non-informative priors $\mathcal{N}(\text{logit}(0.5), 4)$ for the smooth parameters.

**Evaluation.** We use the prediction error on a held-out test set to compare the proposed model to the baseline models. The proposed approach can update parameters online but for the sake of this comparison, we treat the first 50 observations from each individual as training data and the remainder as test. Predictions of the measurements are made under the treatment strategy prescribed in the test set. Since the creatinine levels are measured on average twice a day and treatment decisions are made at the granularity of days, we report the prediction errors for seven days following the end of training. We run 4 randomly initialized chains each for 5,000 iterations with a burn-in of 2,500 iteration and thin of 50 iterations.

To calculate the prediction error, we predict each patient's creatinine levels for seven days based on the individual parameters sampled at every 50 iterations after the burn-in. We compute prediction means individually by averaging each patient's 200 predictions (from the 4 chains each with 50 predictions), and obtain 95% credible intervals individually using the quantiles of the 200 predictions. We calculate the root mean squared error (RMSE) on each individual's prediction mean, and average across the individuals to obtain the overall prediction error and the 95% credible interval.

**Quantitative Results.** In Figure 8, we report the mean prediction errors with the 95% credible intervals for ITR and the three baseline models. ITR outperforms the baseline models significantly after day 3. *Individual* model outperforms *sub-pop* model, and after day 4, it significantly outperforms *pop* model. *Sub-pop* model outperforms *pop* model significantly after day 6. ITR outperforms *sub-pop* model because ITR is more expressive since it allows individual-level heterogeneity and information to be shared across individuals in the same group. On the other hand, ITR also outperforms *inidividual* model because the lack of subgroup structure makes *individual* model statistically less efficient.

**Qualitative Results.** In Figure 9, we present the predictions from ITR and *pop* model for two example patients. Only the last 20 observations (black dots) are plotted for the training set. The red points are creatinine levels that are reserved for the test set. The dashed lines are the predicted baseline progressions and the solid lines are the final predictions of the creatinine levels. Prescriptions of treatments are shown as vertical dashed lines. Treatment response curves are plotted on the right of the trajectory predictions. Ribbons denote the 95% credible intervals.

As an aid for our analysis, we plot a heat map of the renal SOFA (Sequential Organ Failure Assessment) scores (Vincent et al., 1996) above the trajectory predictions. Renal SOFA scores, ranging from 0 to 4, is typically used in the ICU to capture the patient's kidney function. A higher score represents a higher risk for kidney failure. From day 32 to 36, the baseline progression for patient 44 inferred by ITR increases to a greater extent in comparison with the baseline progression inferred by *pop* model. The ITR's inference aligns with the clinical expectation that the patient's creatinine levels will increase without treatment since the patient has a renal SOFA score of 4 during this time period. In comparison to the treatment response curve estimated by *pop* model, the response curve estimated by ITR indicates that the patient is less responsive to CVVHD. The difference between these two estimates could be explained by the fact that although this patient has an overall averaged renal SOFA of 3.9, the average for population is 2.5. For patient 228, who was documented as a Chronic Kidney Disease (CKD) patient, the baseline progression inferred by ITR increases to a smaller extent in comparison with the baseline progression inferred by *pop* model. ITR aligns with the clinical expectation that this patient's creatinine levels will remain stable with RRT treatment since RRT is a modality used to maintain creatinine levels within the normal range in CDK patients with end stage renal disease.

More broadly, we expect the patients to be less responsive to RRT if they have more severe kidney dysfunction. In Figure 10 (a), we plot the estimated treatment response curves associated with the patient's minimum renal SOFA score. A renal SOFA score was calculated each time the creatinine level was measured following the first test that indicated an elevated creatinine level. The curves are normalized over the patient's initial creatinine level at the time when RRT is initiated. We observe that patients with more severely compromised kidney function (as indicated by higher SOFA scores) are more resistant to RRT. In Figure 10(b), we plot the curves for AKI patients, identified by the ICD-9[2] diagnosis code 584, and observe that the AKI patients who also have CKD (identified by ICD-9 codes 585, 585.1–585.6, and 585.9) tend to be more resistant to RRT than those who only have AKI.

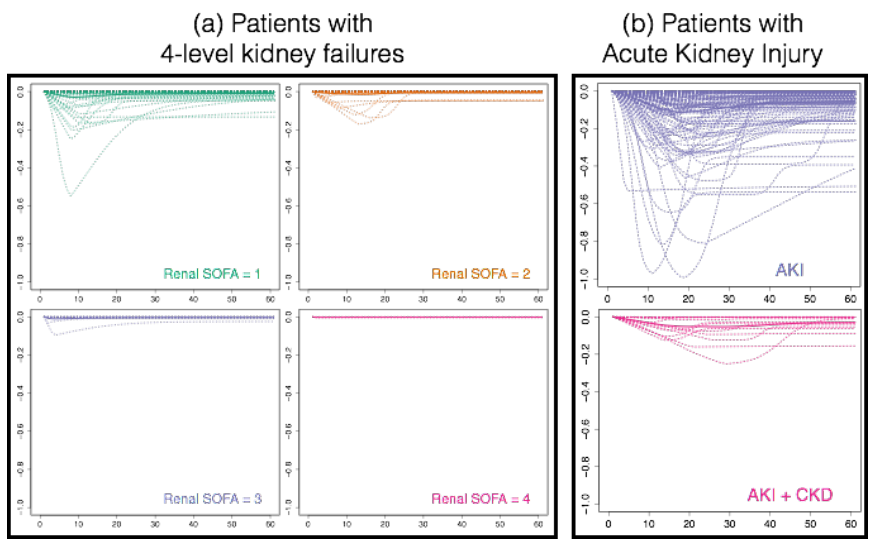

Figure 10: Association of responsiveness to CVVH with kidney disease severity

---

2. http://www.cdc.gov/nchs/icd/icd9.htm

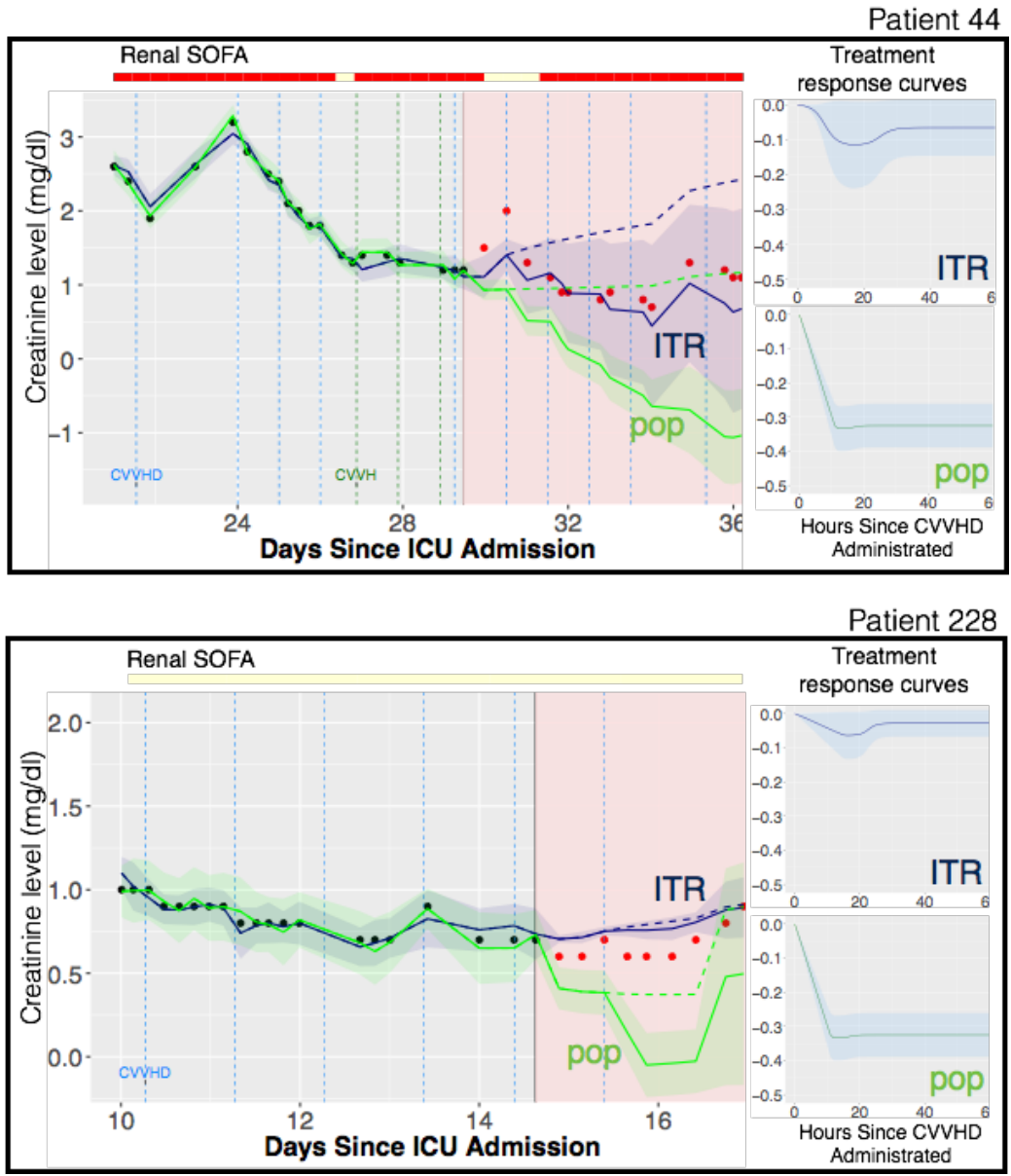

Figure 9: Comparison of ITR vs. *pop* model predictions on two example trajectories for creatinine level. The black points are measurements in the training set and red points are measurements in the test set The dashed lines are the predicted baseline progressions and the solid lines are the final predictions of the creatinine levels. Prescriptions of treatments are shown as vertical dashed lines. Treatment response curves are plotted on the right of the trajectory predictions. Ribbons denote the 95% credible intervals. Heat maps with the colors light yellow (renal SOFA of 0) to red (renal SOFA of 4) are plotted above the trajectory predictions.

### 4.3 Observational Study: Estimating Heterogenous Response Curves to Diuretics

We also evaluate the proposed model on the task of estimating patients' responses to diuretics, particularly Lasix (furosemide), a common treatment to remove excess fluid that has accumulated in critically ill patients after aggressive fluid resuscitation therapies. Fluid overload has been shown to be associated with worse outcomes in many studies. For example, Acheampong and Vincent (2015) show that a positive fluid balance is an independent prognostic factor for mortality in septic patients in the ICU. Although Lasix can be used to treat fluid overload, it should only be used when it is effective (Cerda et al., 2010). Persistent use of diuretics in renal failure or heart failure patients who do not respond to diuretics may delay the initialization of dialysis, another modality that can be used for managing fluid balance. This delay can lead to worse patient outcomes (Cerda et al., 2010). Thus, an individualized assessment tool to predict whether a patient will respond to diuretics can greatly impact the management of fluid status.

**Datasets.** We fit our models on electronic health record data from the MIMIC-II Clinical Database. We estimate the effects of Lasix at four different dose levels: $<= 5$ ml/hr, $<= 10$ ml/hr, $<= 30$ ml/hr, and $> 30$ ml/hr. We use fluid balance since ICU admission as the patient's outcome. We calculate the fluid balance by subtracting the cumulative urine outputs at each time point from the cumulative fluid inputs. We calculate the fluid balance one day before the Lasix is first prescribed. Afterwards, we compute the fluid balance at each time the urine output is updated in the database.

We include the patients who are prescribed Lasix. We exclude the patients' data when other modalities are prescribed for managing the fluid balance, e.g. dialysis. The dataset contains 231 trajectories with a total of $88,578$ observations of fluid balance. Each individual trajectory has an average duration of 51 days. The datasets contains a total of $1,738$ instances of Lasix $<= 5$ ml/hr, $1,146$ of Lasix $<= 10$ ml/hr, 829 of Lasix $<= 30$ ml/hr, and 124 of Lasix $> 30$ ml/hr. Fluid balances were standardized by the population mean of 14.71 and standard deviation of 15.23.

**Experimental setup.** For the fixed-effects component in the baseline progression, we still use $p = 4$ covariates: age, weight, time and 1 for the intercept. We posit a non-informative Normal-inverse-Wishart base distribution NIW$(\mathbf{0}, 1, p+2, I_p)$ for the regression coefficients. We posit a strong prior $\mathcal{N}(\log(0.1^2), 0.3^2)$ on the transformed scale parameters in the exponential Gaussian process kernel and a non-informative prior $\mathcal{N}(\text{logit}(0.5), 4)$ on the transformed smooth parameters.

For the treatment response model, we posit a Gaussian prior on the peak effects with the means chosen by the domain expert. Specifically, these were set as $\mathcal{N}(-1, 1)$, $\mathcal{N}(-2, 1)$, $\mathcal{N}(-3, 1)$, and $\mathcal{N}(-4, 1)$ from low to high dose level of Lasix, respectively. Similarly, we posit a Gaussian prior on the change points with means chosen by the domain expert based on the expected duration over which the treatment takes effect. Specifically, these were set as N(12 hrs, 100 hrs), N(14 hrs, 100 hrs), N(16 hrs, 100 hrs), and N(18 hrs, 100 hrs). We posit non-informative Gaussian priors $\mathcal{N}(\text{logit}(0.5), 4)$ for the two steepness parameters and the ratio of the long-term effect to the peak effect.

For the noise, we posit an Inverse-Gamma prior IG$(1, 1)$ on the variances of the i.i.d. noises. We posit strong priors $\mathcal{N}(\log(0.1^2), 0.3^2)$ on the scale parameters for the time-dependent noises, and non-informative priors $\mathcal{N}(\text{logit}(0.5), 4)$ for the smooth parameters.

**Evaluation.** We use prediction error on a held-out test set to compare the proposed model to the baseline models. We treat the first 100 observations from each individual as training data and the remainder as test data. Predictions of the measurements are made under the treatment strategy prescribed in the test set. Since urine output are measured on average 7 times a day and management decisions are made at the granularity of hours, we report prediction errors for 24 hours following the end of training. Then, we follow the same procedures described in Section 4.2 to evaluate the models.

**Quantitative Results.** In Figure 11(a), we report the mean prediction errors with the 95% credible intervals for ITR and the three baseline models. ITR outperforms the three baseline models significantly. *Pop* model performs the worst across the four models statistically significantly. In Figure 11(b), we demonstrate how the performance of prediction varies across the different models when the number of observations per treatment type varies. *Individual* model performs the worst at the beginning due to the lack of data. As more data are obtained, *individual* model performs better than *pop* and *sub-pop* models, and it eventually performs similarly as ITR. Performance of *pop* model is comparable to ITR when there are few data, but decreases when enough data are given to show the diversity across individuals. It is because *pop* model does not have the flexibility to individualize baseline progressions and treatment response curves like ITR.

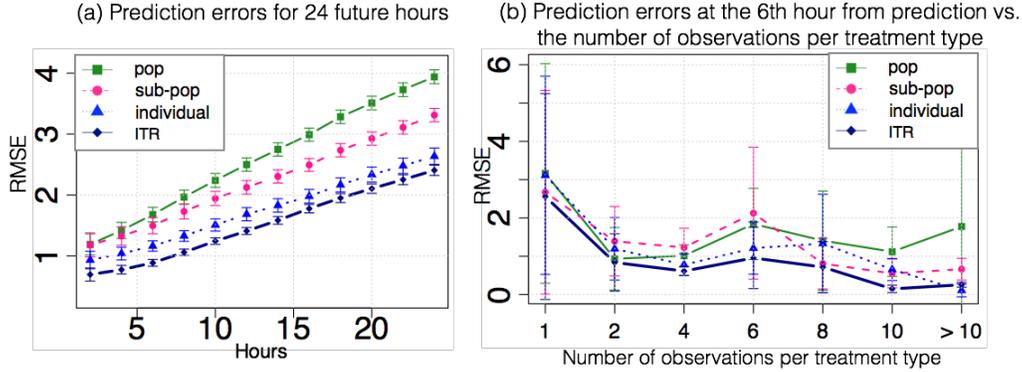

Figure 11: Comparison of ITR vs. baselines prediction errors on fluid balance.

**Qualitative Results.** In Figure 12, we present the predictions from ITR and *pop* model for two example patients. For patient 69, the treatment response curve estimated by ITR indicates that the patient is more responsive to $\leq 5$ml/hr of Lasix in comparison with the response curve estimated by *pop* model. The ITR's estimate aligns with the clinical expectation that the patient will be responsive to low dose levels of Lasix since the patient has an overall averaged renal SOFA score of 0. The reason *Pop* model estimates a less responsive curve for low dose level of Lasix could be explained by the fact that the average renal SOFA for population is 1.2. For patient 30, ITR can not predict the final outcomes accurately because the creatinine level increases unexpectedly when the patient develops AKI during the time of prediction. The renal SOFA score also has a sudden increase from 1 to 4 within a day. However, in contrast with the estimates from *pop* model, the high stabilized baseline progression and the resistant response curve for high dose level of Lasix (between 10ml/hr and 30ml/hr) estimated by ITR potentially indicate the decreased kidney function.

More broadly, response to diuretics can indicate lesser severity of kidney disease (Cerda et al., 2010). In Figure 13(a), we plot the estimated treatment response curves associated with the patient's minimum renal SOFA score. The curves are normalized over the patient's initial fluid balance at the time when Lasix was prescribed. We observe that the effect of diuretics on fluid balance tends to decrease when the patient's renal function declines. In Figure 13(b), we also see that the patients who have both AKI and CKD tend to be more resistant to Lasix than those who only have AKI. In addition, dialysis may be initiated as an alternative modality to remove the the excess fluid if the patients are observed to be resistant to diuretics. To validate this, in Figure 13(c) we plot the estimated treatment response curves associated with an indicator whether the patient is prescribed dialysis in the future or not. As we expect, the patients who are more responsive to diuretics do not need dialysis for further treatments.

## 5. Conclusion

In this paper, we have developed a novel Bayesian nonparametric method for estimating treatment response curves from sparse observational time series. We leverage hierarchical priors that allow individual-specific estimates while borrowing information across individuals. Notably, we maintain the full posterior rather than just point estimates. We demonstrate significant gains in performance for modeling creatinine and effects of treatments used for managing kidney function, as well as effects of diuretics used for managing fluid balance. As future work, we plan to evaluate these models on other subpopulations with MIMIC and test sensitivity to different modeling choices. Access to accurate models for estimating treatments responses at the individual are critical for designing new personalized treatments.

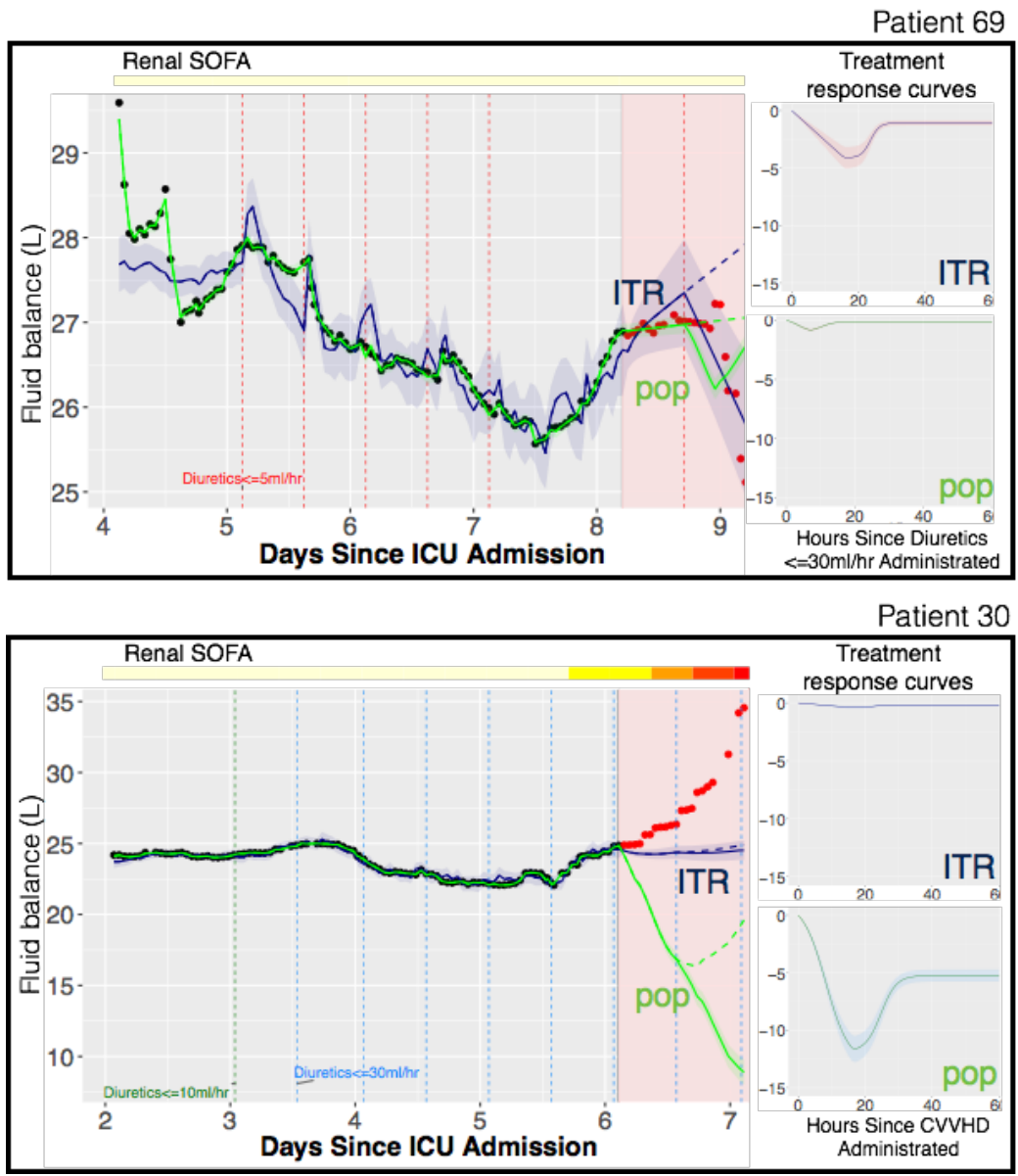

Figure 12: Comparison of ITR vs. *pop* predictions on two example trajectories for fluid balance. The black and red points are training and predicting measurements, respectively. The dashed lines are the predicted baseline progressions and the solid lines are the final predictions of the creatinine levels. Prescriptions of treatments are shown as vertical dashed lines. Treatment response curves are plotted on the right of the trajectory predictions. Ribbons denote the 95% credible intervals for the prediction. Heat maps with the colors light yellow (renal SOFA of 0) to red (renal SOFA of 4) are plotted above the trajectory predictions.

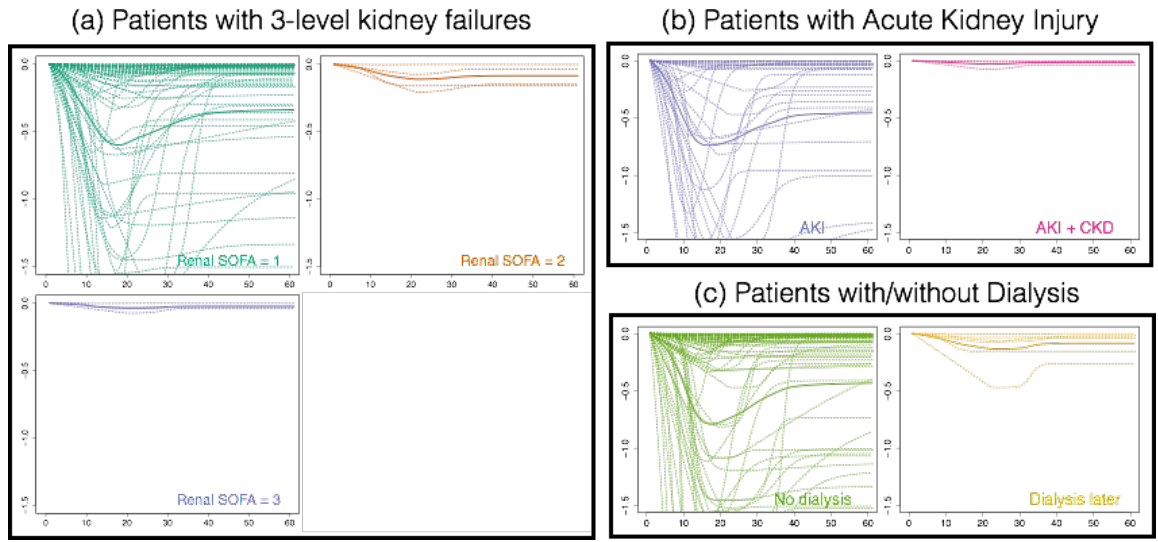

Figure 13: Association of patients' responsiveness to Lasix $\leq 30$ml/hr with patients' kidney disease severity (No patients in the dataset start with renal SOFA score 4).

## Appendix A. Transformation of Constrained Variables

The treatment-response curves were characterized using a parametric form containing constrained variables (e.g., $\alpha_2, \alpha_3 \in (0,1)$). To simplify inference, we transform the support of these variables such that they live in the real space $\mathbb{R}$ and posit (Gaussian) priors on these transformed variables. Given a random variable $X \in \mathbb{R}^d$ with continuous probability density function $f_X(x)$ and support $\mathcal{X} = \text{supp}(f_X(x))$, we can define a random variable $Y \in \mathbb{R}^d$ such that $Y = T(X)$ with support $\mathcal{Y} = \text{supp}(f_Y(y))$ and a one-to-one differentiable function $T : \mathcal{X} \to \mathcal{Y}$. Then based on Olive (2014), $Y$ has the probability density function

$$f_Y(y) = f_X(T^{-1}(y))|\det J_{T^{-1}}(y)|,$$

where the adjustment term is the absolute determinant of the Jacobian:

$$J_{T^{-1}}(Y) = \begin{pmatrix} \frac{\partial T_1^{-1}}{\partial y_1} & \cdots & \frac{\partial T_1^{-1}}{\partial y_d} \\ \vdots & & \vdots \\ \frac{\partial T_d^{-1}}{\partial y_1} & \cdots & \frac{\partial T_d^{-1}}{\partial y_d} \end{pmatrix}$$

Let us first consider the univariate variable $\alpha_2 \in (0,1)$ from the $g$ function we defined in Section 2.2. We transform it to be $\alpha_2' = \text{logit}(\alpha_2)$, and posit a Gaussian prior on it. That is, $\alpha_2' \sim \mathcal{N}(\alpha_2'; \mu_{\alpha_2'}, \sigma_{\alpha_2'}^2)$. The Jacobian adjustment is calculated as $|\det J(\text{logit}(\alpha_2))| = 1/\alpha_2(1-\alpha_2)$. Thus we get the probability density function

$$p(\alpha_2) = \mathcal{N}(\text{logit}(\alpha_2); \mu_{\alpha_2'}, \sigma_{\alpha_2'}^2)/\alpha_2(1-\alpha_2).$$

Now let us consider the multivariate variable $\boldsymbol{\phi} = \{\alpha_1, \alpha_2, \alpha_3, \gamma, b : \alpha_1 \in \mathbb{R}, \alpha_2, \alpha_3 \in (0,1), \gamma \in \mathbb{R}, b/g(\gamma) \in (0,1)\}$ in the $g$ function. We define a transformation

$$\boldsymbol{\phi}' = T^{-1}(\boldsymbol{\phi}) = \{\alpha_1, \text{logit}(\alpha_2), \text{logit}(\alpha_3), \gamma, \text{logit}(b/g(\gamma))\},$$

where $g(\gamma) = \alpha_1(\exp(\alpha_2\gamma/2) - 1)/(\exp(\alpha_2\gamma/2) + 1)$. Since the support of $\boldsymbol{\phi}'$ is $\mathbb{R}^d$, we can posit a diagonal Gaussian prior $\boldsymbol{\phi}' \sim \mathcal{N}(\boldsymbol{\phi}'; \boldsymbol{\mu}_{\boldsymbol{\phi}'}, \boldsymbol{D}_{\boldsymbol{\phi}'})$, and calculate the Jacobian

$$J_{T^{-1}}(\boldsymbol{\phi}) = \begin{pmatrix} 1 & 0 & 0 & 0 & 0 \\ 0 & 1/\alpha_2(1-\alpha_2) & 0 & 0 & 0 \\ 0 & 0 & 1/\alpha_3(1-\alpha_3) & 0 & 0 \\ 0 & 0 & 0 & 1 & 0 \\ \frac{b\zeta}{\alpha_1} & \frac{b\gamma\zeta\exp(\alpha_2\gamma/2)}{\exp(\alpha_2\gamma)-1} & 0 & \frac{b\alpha_2\zeta\exp(\alpha_2\gamma/2)}{\exp(\alpha_2\gamma)-1} & \zeta \end{pmatrix}.$$

Here, $\zeta = g(\gamma)/b(g(\gamma) - b)$. Thus we obtain the adjustment $|\det J_{T^{-1}}(\boldsymbol{\phi})| = |\zeta|/\alpha_2\alpha_3(1-\alpha_2)(1-\alpha_3)$ and reach at the probability density function

$$p(\boldsymbol{\phi}) = \mathcal{N}(T^{-1}(\boldsymbol{\phi}); \boldsymbol{\mu}_{\boldsymbol{\phi}'}, \boldsymbol{D}_{\boldsymbol{\phi}'})|\zeta|/\alpha_2\alpha_3(1-\alpha_2)(1-\alpha_3).$$

We also have constrained parameters in the exponential kernels: $\sigma^2 \in \mathbb{R}^+$ and $\rho \in (0,1)$ (to be more precise, $\sigma_{ui}^2, \rho_{ui}$ and $\sigma_{\epsilon_d'}^2, \rho_{\epsilon_d'}$). We again define transformations $\sigma'^2 = \log(\sigma^2)$ and $\rho' = \text{logit}(\rho)$ and posit Gaussian priors on them. Thereafter, we get the densities $p(\sigma^2) = \mathcal{N}(\log(\sigma^2); \mu_{\sigma'}, \sigma_{\sigma'}^2)/\sigma^2$ and $p(\rho) = \mathcal{N}(\text{logit}(\rho); \mu_{\rho'}, \sigma_{\rho'}^2)/\rho(1-\rho)$ respectively.

## Appendix B. Posterior inference for the individualized treatment response model

### B.1 Blokced Gibbs Sampler for the DPM

We first summarize the blocked Gibbs sampler for general DPMs, and then apply it specifically to $\boldsymbol{\varphi}_i$ and $\boldsymbol{\phi}_i$ in our model.

Given a sufficiently large $K$, the mixture component parameters $\boldsymbol{\theta}^* = \{\theta_1^*, ..., \theta_K^*\}$, the stick breaking variables $\boldsymbol{V} = \{V_1, ..., V_{K-1}, V_K = 1\}$ and the component indicators $\boldsymbol{Z} = \{Z_1, ..., Z_N\}$ for the $N$ observations $\boldsymbol{o} = \{o_1, ..., o_N\}$, the truncated stick-breaking representation of DPM is written as follows.

$$p(o_n|\pi_k, \theta_k^*) = \sum_{k=1}^{K} \pi_k p(o_n|\theta_k^*) \tag{13}$$

$$\pi_k = V_k \prod_{j<k}(1 - V_j), \text{ for } k = 1, ..., K$$

$$V_k \sim \text{Beta}(1, M), \text{ for } k = 1, ..., K-1$$

Then the blocked Gibbs sampler is formulated by the following steps.

1. Independently sample $\theta_k^*$ from $p(\theta_k^*|\boldsymbol{Z}, \boldsymbol{V}, \boldsymbol{o}) \propto G_0(\theta_k^*) \prod_{n=1}^{N} p(o_n|\theta_k^*) \mathbf{1}_{\{Z_n = k\}}$;

2. Independently sample $v_k$ from $p(V_k|\boldsymbol{Z}, \boldsymbol{\theta}, \boldsymbol{o}) = \text{Beta}(1 + n_k, M + \sum_{j=k+1}^{K} n_j)$, where $n_j$ is the number of observations in cluster $j$;

3. Independently sample $z_n$ from $p(Z_n = k|\boldsymbol{V}, \boldsymbol{\theta}, \boldsymbol{o}) = \pi_k p(o_n|\theta_k)$, where $\pi_k = V_k \prod_{j<k}(1 - V_j)$.

Note that step 1 can be derived in closed form if the base distribution $G_0$ is chosen to be conjugate—a choice we make in (8). Now let us specify the samplers for the DPM parameters in our model.

We first describe the steps of sampling DP mixtures for $\boldsymbol{\varphi}_i$'s. Suppose $K_1$ is the truncation level we assume for the baseline progression. Denote the mixture component hyperparameters as $\boldsymbol{\theta}_\varphi^* = \{\boldsymbol{\beta}_b^*, \Sigma_b^*, \boldsymbol{\mu}_{\sigma_u'}^*, \boldsymbol{\mu}_{\rho_u'}^*\}$, where $\boldsymbol{\beta}_b^* = \{\beta_{b_1}^*, ..., \beta_{b_{K_1}}^*\}$, $\Sigma_b^* = \{\Sigma_{b_1}^*, ..., \Sigma_{b_{K_1}}^*\}$, $\boldsymbol{\mu}_{\sigma_u'}^* = \{\mu_{\sigma_{u_1}'}^*, ..., \mu_{\sigma_{u_{K_1}}'}^*\}$, $\boldsymbol{\mu}_{\rho_u'}^* = \{\mu_{\rho_{u_1}'}^*, ..., \mu_{\rho_{u_{K_1}}'}^*\}$. Further, the stick breaking variables $\boldsymbol{V}_\varphi = \{V_{\varphi_1}, ..., V_{\varphi_{K_1-1}}, V_{\varphi_{K_1}} = 1\}$ and the component indicators $\boldsymbol{Z}_\varphi = \{Z_{\varphi_1}, ..., Z_{\varphi_I}\}$ for the parameters $\boldsymbol{\varphi} = \{\varphi_1, .., \varphi_I\}$.

1. Independently sample $\boldsymbol{\beta}_{b_k}^*$, $\Sigma_{b_k}^*$ from

$$p(\boldsymbol{\beta}_{b_k}^*, \Sigma_{b_k}^*|\boldsymbol{Z}_\varphi, \boldsymbol{V}_\varphi, \boldsymbol{\varphi}) = \text{NIW}(\boldsymbol{\beta}_{b_k}^*, \Sigma_{b_k}^*|\boldsymbol{m}_k, \kappa_k, \nu_k, \boldsymbol{S}_k)$$

$$\boldsymbol{m}_k = \frac{\kappa_0 \beta_0 + \sum_{i=1}^{I} \boldsymbol{\beta}_i \mathbf{1}_{\{Z_{\varphi_i} = k\}}}{\kappa_k}$$

$$\kappa_k = \kappa_0 + \sum_{i=1}^{I} \mathbf{1}_{\{Z_{\varphi_i} = k\}}$$

$$\nu_k = \nu_0 + \sum_{i=1}^{I} \mathbf{1}_{\{Z_{\varphi_i} = k\}}$$

$$\boldsymbol{S}_k = \boldsymbol{S}_0 + \sum_{i=1}^{I} \boldsymbol{\beta}_i \boldsymbol{\beta}_i^T \mathbf{1}_{\{Z_{\varphi_i} = k\}} + \kappa_0 \beta_0 \beta_0^T - \kappa_k \boldsymbol{m}_k \boldsymbol{m}_k^T;$$

2. Independently sample $\boldsymbol{\mu}_{\sigma_{u_k}'}^*$ from

$$p(\boldsymbol{\mu}_{\sigma_{u_k}'}^*|\boldsymbol{Z}_\varphi, \boldsymbol{V}_\varphi, \boldsymbol{\varphi}) = \mathcal{N}(\boldsymbol{\mu}_{\sigma_{u_k}'}^*; m_{\sigma_u'}, s_{\sigma_u'})$$

$$m_{\sigma'} = \frac{\sigma_{\sigma_{u0}'}^2 \mu_{\sigma_0'} + \sigma_{\sigma_0'}^2 \sum_{i=1}^{I} \log(\sigma_{ui}^2) \mathbf{1}_{\{Z_{\varphi_i} = k\}}}{\sigma_{\sigma_{u0}'}^2 + \sum_{i=1}^{I} \sigma_{\sigma_0'}^2 \mathbf{1}_{\{Z_{\varphi_i} = k\}}}$$

$$s_{\sigma_u'} = \frac{\sigma_{\sigma_{u0}'}^2 \sigma_{\sigma_0'}^2}{\sigma_{\sigma_{u0}'}^2 + \sum_{i=1}^{I} \sigma_{\sigma_0'}^2 \mathbf{1}_{\{Z_{\varphi_i} = k\}}}$$

3. Independently sample $\boldsymbol{\mu}^*_{\rho'_{u_k}}$ from

$$p(\boldsymbol{\mu}^*_{\rho'_{u_k}}|\boldsymbol{Z}_\varphi, \boldsymbol{V}_\varphi, \boldsymbol{\varphi}) = \mathcal{N}(\boldsymbol{\mu}^*_{\rho'_{u_k}}; m_{\rho'_u}, s_{\rho'_u})$$

$$m_{\rho'} = \frac{\sigma^2_{\rho'_{u0}}\mu_{\rho'_0} + \sigma^2_{\rho'_0}\sum_{i=1}^I \text{logit}(\rho_{ui})\mathbf{1}_{\{Z_{\varphi_i}=k\}}}{\sigma^2_{\rho'_{u0}} + \sum_{i=1}^I \sigma^2_{\rho'_0}\mathbf{1}_{\{Z_{\varphi_i}=k\}}}$$

$$s_{\rho'_u} = \frac{\sigma^2_{\rho'_{u0}}\sigma^2_{\rho'_0}}{\sigma^2_{\rho'_{u0}} + \sum_{i=1}^I \sigma^2_{\rho'_0}\mathbf{1}_{\{Z_{\varphi_i}=k\}}}$$

4. Independently sample $V_{\varphi_k}$ from

$$p(V_{\varphi_k}|\boldsymbol{Z}_\varphi, \boldsymbol{\theta}^*_\varphi, \boldsymbol{\varphi}) = \text{Beta}(1 + n_{1k}, M_1 + \sum_{j=k+1}^{K_1} n_{1j}),$$

where $n_{1j}$ is the number of $\boldsymbol{\varphi}_i$'s that were assigned to cluster $j$;

5. Independently sample $Z_{\varphi_i}$ from

$$p(Z_{\varphi_i} = k|\boldsymbol{V}_\varphi, \boldsymbol{\theta}^*_\varphi, \boldsymbol{\varphi})$$
$$= \omega_{1k}\mathcal{N}(\boldsymbol{\beta}_i; \boldsymbol{\beta}^*_{b_k}, \Sigma^*_{b_k})\mathcal{N}(\log(\sigma^2_{u_i}); \mu^*_{\sigma'_{u_k}}, \sigma^2_{\sigma'_{u0}})\mathcal{N}(\text{logit}(\rho_{u_i}); \mu^*_{\rho'_{u_k}}, \sigma^2_{\rho'_{u0}})/\sigma^2_{u_i}(1-\rho_{u_i})^2,$$

where $\omega_{1k} = V_{\varphi_k}\prod_{j<k}(1-V_{\varphi_j})$.

Now we describe the steps of sampling DP mixtures for $\boldsymbol{\phi}_{id}$'s. Let $K_{2d}$ be the truncation level assumed for the DPM prior on the $d$th treatment-response for $(d = 1, ..., D)$. Denote the mixture component hyperparameters as $\boldsymbol{\theta}^*_{\phi_d} = \{\boldsymbol{\mu}^*_{\phi'_d}\}$, where $\boldsymbol{\mu}^*_{\phi'_d} = \{\boldsymbol{\mu}^*_{\phi'_{d1}}, ..., \boldsymbol{\mu}^*_{\phi'_{dK_{2d}}}\}$. Further, the stick breaking variables $\boldsymbol{V}_{\phi_d} = \{V_{\phi_{d1}}, ..., V_{\phi_{dK_{2d}-1}}, V_{\phi_{dK_{2d}}} = 1\}$ and the component indicators $\boldsymbol{Z}_{\phi_d} = \{Z_{\phi_{1d}}, ..., Z_{\phi_{Id}}\}$ for the parameters $\boldsymbol{\phi}_d = \{\boldsymbol{\phi}_{1d}, ..., \boldsymbol{\phi}_{Id}\}$.

6. Independently sample $\boldsymbol{\mu}^*_{\phi'_d}$ from

$$p(\boldsymbol{\mu}^*_{\phi'_d}|\boldsymbol{Z}_{\phi_d}, \boldsymbol{V}_{\phi_d}, \boldsymbol{\phi}_d) = \mathcal{N}(\boldsymbol{\mu}^*_{\phi'_d}; \boldsymbol{m}_{\phi'_d}, \boldsymbol{S}_{\phi'_d})$$

$$\boldsymbol{S}_{\phi'_d} = (\boldsymbol{D}^{-1}_{0_d} + \boldsymbol{D}^{-1}_{\phi'_0}\sum_{i=1}^I \mathbf{1}_{\{Z_{\phi_{id}}=k\}})^{-1}$$

$$\boldsymbol{m}_{\phi'_d} = \boldsymbol{S}_{\phi'_d}(\boldsymbol{D}^{-1}_{0_d}\boldsymbol{\mu}_{0_d} + \boldsymbol{D}^{-1}_{\phi'_0}\sum_{i=1}^I T^{-1}(\boldsymbol{\phi}_{id})\mathbf{1}_{\{Z_{\phi_{id}}=k\}}),$$

where $T^{-1}(\boldsymbol{\phi}_{id}) = \{\alpha_{1_{id}}, \text{logit}(\alpha_{2_{id}}), \text{logit}(\alpha_{3_{id}}), \gamma_{id}, \text{logit}(b_{id}/g(\gamma_{id}))\}$;

7. Independently sample $V_{\phi_{dk}}$ from

$$p(V_{\phi_{dk}}|\boldsymbol{Z}_{\phi_d}, \boldsymbol{\theta}^*_{\phi_d}, \boldsymbol{\phi}_d) = \text{Beta}(1 + n_{2dk}, M_{2d} + \sum_{j=k+1}^{K_{2d}} n_{2dj}),$$

where $n_{2dj}$ is the number of $\boldsymbol{\phi}_{id}$'s that were assigned to cluster $j$;

8. Independently sample $Z_{\phi_{id}}$ from

$$p(Z_{\phi_{id}} = k|\boldsymbol{V}_{\phi_d}, \boldsymbol{\theta}^*_{\phi_d}, \boldsymbol{\phi}_d) = \omega_{2dk}\mathcal{N}(T^{-1}(\boldsymbol{\phi}_{id}); \boldsymbol{\mu}^*_{\phi'_{dk}}, \boldsymbol{D}_{\phi'_0})|\zeta_{id}|/(1-\alpha_{2_{id}})^4,$$

where $\omega_{2dk} = V_{\phi_{dk}}\prod_{j<k}(1-V_{\phi_{dj}})$.

## B.2 Gibbs Sampler for the Variables with Conjugate Priors

9. Independently sample $\boldsymbol{\beta}_i$ from

$$p(\boldsymbol{\beta}_i|\boldsymbol{\theta}_\varphi^*, Z_{\varphi_i}, \sigma_{\epsilon_i}^2, \boldsymbol{b}_i, \boldsymbol{f}_i) = \mathcal{N}(\boldsymbol{\beta}_i; \boldsymbol{m}_{b_i}, \boldsymbol{S}_{b_i})$$
$$\boldsymbol{S}_{b_i} = (\Sigma_{bZ_{\varphi_i}}^{*-1} + \sigma_{\epsilon_i}^{-2} \boldsymbol{X}_i \boldsymbol{X}_i^T)^{-1}$$
$$\boldsymbol{m}_{b_i} = \boldsymbol{S}_{b_i}(\sigma_{\epsilon_i}^{-2} \boldsymbol{X}_i \boldsymbol{Y}_{b_i} + \Sigma_{bZ_{\varphi_i}}^{*-1} \boldsymbol{\mu}_{bZ_{\varphi_i}}^*),$$

where $\boldsymbol{X}_i = \{\boldsymbol{X}_{i1}, ..., \boldsymbol{X}_{iJ_i}\}^T$ is a $J_i \times p$ matrix, $\boldsymbol{Y}_{b_i} = \boldsymbol{Y}_i - \boldsymbol{b}_i - \boldsymbol{f}_i$ is a $1 \times J_i$ vector and $\boldsymbol{f}_i$ is defined in Eq. (4). $\Sigma_{bZ_{\varphi_i}}^{*-1}$ and $\boldsymbol{\mu}_{bZ_{\varphi_i}}^*$ are sampled in Step 1.

10. Independently sample $\boldsymbol{b}_i$ from

$$p(\boldsymbol{b}_i|\boldsymbol{\beta}_i, \boldsymbol{\varphi}_i, \sigma_{\epsilon_i}^2, \boldsymbol{f}_i) = \mathcal{N}(\boldsymbol{b}_i; \boldsymbol{m}_{u_i}, \boldsymbol{S}_{u_i})$$
$$\boldsymbol{S}_{u_i} = \left(\mathcal{K}_u^{-1}(t_{ij}, t_{ij'}; \boldsymbol{\varphi}_i) + \sigma_{\epsilon_i}^{-2} \boldsymbol{I}_{J_i}\right)^{-1}$$
$$\boldsymbol{m}_{u_i} = \sigma_{\epsilon_i}^{-2} \boldsymbol{S}_{u_i} \boldsymbol{Y}_{u_i},$$

where $\boldsymbol{Y}_{u_i} = \boldsymbol{Y}_i - \boldsymbol{X}_i \boldsymbol{\beta}_i - \boldsymbol{f}_i$ is a $1 \times J_i$ vector.

11. Independently sample $\sigma_{\epsilon_i}^2$ from

$$p(\sigma_{\epsilon_i}^2|\boldsymbol{\beta}_i, \boldsymbol{b}_i, \boldsymbol{f}_i') = \text{IG}(s_\epsilon + J_i/2, \nu + \boldsymbol{Y}_{e_i} \boldsymbol{Y}_{e_i}^T/2),$$

where $\boldsymbol{Y}_{e_i} = \boldsymbol{Y}_i - \boldsymbol{X}_i \boldsymbol{\beta}_i - \boldsymbol{b}_i - \boldsymbol{f}_i'$ is a $1 \times J_i$ vector and the auxiliary variable $\boldsymbol{f}_i'$ is sampled from

$$p(\boldsymbol{f}_i'|\boldsymbol{\beta}_i, \boldsymbol{\phi}_i, \boldsymbol{\sigma}_f^2, \boldsymbol{\rho}_f^2, \sigma_{\epsilon_i}^2, \boldsymbol{b}_i) = \mathcal{N}(\boldsymbol{f}_i'; \boldsymbol{m}_{f_i}, \boldsymbol{S}_{f_i})$$
$$\boldsymbol{S}_{f_i} = \left(\mathcal{K}_f^{-1}(t_{ij}, t_{ij'}; \boldsymbol{\sigma}_f^2, \boldsymbol{\rho}_f^2) + \sigma_{\epsilon_i}^{-2} \boldsymbol{I}_{J_i}\right)^{-1}$$
$$\boldsymbol{m}_{f_i} = \boldsymbol{S}_{f_i}\left(\sigma_{\epsilon_i}^{-2} \boldsymbol{Y}_{f_i} + \mathcal{K}_f^{-1}(t_{ij}, t_{ij'}; \boldsymbol{\sigma}_f^2, \boldsymbol{\rho}_f^2) m(\boldsymbol{t}_i; \boldsymbol{\phi}_i)\right),$$

where $\boldsymbol{Y}_{f_i} = \boldsymbol{Y}_i - \boldsymbol{X}_i \boldsymbol{\beta}_i - \boldsymbol{b}_i$ is a $1 \times J_i$ vector.

12. Sample $M_1$ from

$$p(M_1|\eta_1, k_1) \sim \frac{c_1 + k_1 - 1}{c_1 + k_1 - 1 + I(d_1 - \log(\eta_1))} \text{Gamma}(c_1 + k_1, d_1 - \log(\eta_1))$$
$$+ \frac{I(d_1 - \log(\eta_1))}{c_1 + k_1 - 1 + I(d_1 - \log(\eta_1))} \text{Gamma}(c_1 + k_1 - 1, d_1 - \log(\eta_1)),$$

where the auxiliary variable $\eta_1 \sim \text{Beta}(M_1 + 1, I)$, the prior for $M_1$ is $\text{Gamma}(c_1, d_1)$, and $k_1$ is the current cluster number for $\boldsymbol{\varphi}_i$'s.

13. Independently sample $M_{2d}$ from

$$p(M_{2d}|\eta_{2d}, k_{2d}) \sim \frac{c_{2d} + k_{2d} - 1}{c_{2d} + k_{2d} - 1 + I(d_{2d} - \log(\eta_{2d}))} \text{Gamma}(c_{2d} + k_{2d}, d_{2d} - \log(\eta_{2d}))$$
$$+ \frac{I(d_{2d} - \log(\eta_{2d}))}{c_{2d} + k_{2d} - 1 + I(d_{2d} - \log(\eta_{2d}))} \text{Gamma}(c_{2d} + k_{2d} - 1, d_{2d} - \log(\eta_{2d})),$$

where the auxiliary variable $\eta_{2d} \sim \text{Beta}(M_{2d} + 1, I)$, the prior for $M_{2d}$ is $\text{Gamma}(c_{2d}, d_{2d})$, and $k_{2d}$ is the current cluster number for $\boldsymbol{\phi}_{id}$'s.

### B.3 Metropolis-Hastings Sampler in the Non-Conjugate Case

We use blocked Metropolis-Hastings to sample the remaining parameters i.e., parameters for which we cannot obtain the conditional distributions in closed-form: $\sigma_{u_i}^2$, $\rho_{u_i}$, $\sigma_{\epsilon_d'}^2$, $\rho_{\epsilon_d'}$, and $\phi_{id}$'. Specifically, for a variable $x$, we propose a candidate value $x^{\text{cand}}$ from a proposal distribution $p(x^{\text{cand}}|x)$ and accept the candidate with probability

$$\min\{1, \frac{\pi(x^{\text{cand}})p(x|x^{\text{cand}})}{\pi(x)p(x^{\text{cand}}|x)}\},$$

where $\pi(\cdot)$ is the full joint posterior defined in Eq. (12). Below, we choose different proposal distributions for $x$ for the following three different types of support.

- For $x \in \mathbb{R}$, we propose new sampler from $\mathcal{N}(x, 0.3^2)$, which is a symmetric proposal distribution.

- For $x \in \mathbb{R}^+$, we propose new sampler from $\mathcal{N}(x, 0.3^2)/\Phi(x, 0.3^2)$, where $\Phi$ is the CDF of the normal distribution. This is not a symmetric proposal distribution.

- For $x \in (0, 1)$, we propose a new sampler from $\mathcal{N}(x, 0.15^2)$ and "reflect" it by 0 or 1 to make it fall back in $(0, 1)$. This is so-called "reflected normal", and the reflection can be done multiple times if needed. It is still a symmetric proposal distribution.

We experimented with a few different choice of values for the variance parameter in the proposal distribution. The values selected above yielded reasonable acceptance rates in the range of $0.14 - 0.22$.

In detail, the sampling for the remainder of the parameters proceeds as follows.

14. Propose $\sigma_{u_i}^{2\text{cand}} \sim \mathcal{N}(\sigma_{u_i}^2, 0.3^2)/\Phi(\sigma_{u_i}^2, 0.3^2)$, $\rho_{u_i}^{\text{cand}} \sim \mathcal{N}(\rho_{u_i}, 0.15^2)$ an reflect $\rho_{u_i}^{\text{cand}}$ into $(0, 1)$. We accept the proposal with probability of $\min\{1, \frac{\pi(\sigma_{u_i}^{2\text{cand}}, \rho_{u_i}^{\text{cand}})\Phi(\sigma_{u_i}^2)}{\pi(\sigma_{u_i}^2, \rho_{u_i})\Phi(\sigma_{u_i}^{2\text{cand}})}\}$, where $\pi(\sigma_{u_i}^{2\cdot}, \rho_{u_i}^{\cdot})$ is

$$\mathcal{N}(Y_{u_i}; 0, \mathcal{K}_u(\beta_i, \sigma_{u_i}^{2\cdot}, \rho_{u_i}^{\cdot}) + \sigma_{\epsilon_i}^2 I_{J_i})\mathcal{N}(\log(\sigma_{u_i}^{2\cdot}); \mu_{\sigma_{ui}'}, \sigma_{\sigma_{u0}'}^2)\mathcal{N}(\text{logit}(\rho_{u_i}^{\cdot}); \mu_{\rho_{ui}'}, \sigma_{\rho_{u0}'}^2)/\sigma_{u_i}^{2\cdot}(1-\rho_{u_i}^{\cdot})^2.$$

15. Propose $\sigma_{\epsilon'}^{2\text{cand}} \sim \mathcal{N}(\sigma_{\epsilon'}^2, 0.3^2 I_D)/\Phi(\sigma_{\epsilon'}^2, 0.3^2 I_D)$ and $\rho_{\epsilon'}^{\text{cand}} \sim \mathcal{N}(\rho_{\epsilon'}, 0.15^2 I_D)$ and reflect $\rho_{\epsilon'}^{\text{cand}}$ into $(0, 1)^D$. We accept the proposal with probability of $\min\{1, \frac{\pi(\sigma_{\epsilon'}^{2\text{cand}}, \rho_{\epsilon'}^{\text{cand}})\Phi(\sigma_{\epsilon'}^2, 0.3^2 I_D)}{\pi(\sigma_{\epsilon'}^2, \rho_{\epsilon'})\Phi(\sigma_{\epsilon'}^{2\text{cand}}, 0.3^2 I_D)}\}$, where $\pi(\sigma_{\epsilon'}^{2\cdot}, \rho_{\epsilon'}^{\cdot})$ is

$$\prod_{i=1}^I \mathcal{N}(Y_{e_i}; m(t_i, \phi_i), \mathcal{K}_{\epsilon'}(\sigma_{\epsilon'}^{2\cdot}, \rho_{\epsilon'}^{\cdot}) + \sigma_{\epsilon_i}^2 I_{J_i}) \prod_{d=1}^D \mathcal{N}(\log(\sigma_{\epsilon_d'}^{2\cdot}); \mu_{\epsilon_1'}, \sigma_{\epsilon_1'}^2)$$
$$\mathcal{N}(\text{logit}(\rho_{\epsilon_d'}^{\cdot}); \mu_{\epsilon_2'}, \sigma_{\epsilon_2'}^2)/\sigma_{\epsilon_d'}^{2\cdot}(1 - \rho_{\epsilon_d'}^{\cdot})^2.$$

16. Propose $\{\alpha_{1id}^{\text{cand}}, \alpha_{2id}^{\text{cand}}, \alpha_{3id}^{\text{cand}}, \gamma_{id}^{\text{cand}}, b_{id}^{\text{cand}}/g(\gamma_{id}^{\text{cand}})\} \sim \mathcal{N}(\{\alpha_{1id}, \alpha_{2id}, \alpha_{3id}, \gamma_{id}, b_{id}/g(\gamma_{id})\},$ Diag$(0.3^2, 0.15^2, 0.15^2, 0.3^2, 0.15^2))$, $\alpha_{2id}^{\text{cand}} \sim \mathcal{N}(\alpha_{1id}, 0.3^2)$ and reflect $\alpha_{2id}^{\text{cand}}, \alpha_{3id}^{\text{cand}}$ and $b_{id}^{\text{cand}}/g(\gamma_{id}^{\text{cand}})$ into $(0, 1)$ individually. We accept the proposal with probability of $\min\{1, \frac{\pi(\phi_i^{\text{cand}})}{\pi(\phi_i)}\}$, where $\pi(\phi_i^{\cdot})$ is

$$\mathcal{N}(Y_{e_i}; m(t_i, \phi_i^{\cdot}), \mathcal{K}_f(\sigma_{\epsilon'}^2, \rho_{\epsilon'}) + \sigma_{\epsilon_i}^2 I_{J_i}) \prod_{d=1}^D \mathcal{N}(T^{-1}(\phi_{id}^{\cdot}); \mu_{\phi_{id}'}, D_{\phi_0'})|\zeta_{id}^{\cdot}|/(1 - \alpha_{2id}^{\cdot})^4.$$